\documentclass[journal]{IEEEtran}

\usepackage{amsmath,amssymb,amsfonts}
\usepackage{algorithmic}
\usepackage{graphicx}
\usepackage{textcomp}

\usepackage[utf8]{inputenc}

\usepackage{hyperref}
\usepackage[style=ieee,sorting=none]{biblatex}
\usepackage{array}
\usepackage{tikz}
\usepackage[export]{adjustbox}

\newcommand{\Ie}{I.e\@.}
\newcommand{\ie}{i.e\@.}

\newcommand{\Eg}{E.g\@.}
\newcommand{\eg}{e.g\@.}

\newcommand{\vs}{vs\@.}
\newcommand{\etal}{\textit{et al\@.}}

\newcommand{\theader}[1]{\textit{#1}}
\newcommand{\SectionRef}[1]{\hyperref[#1]{Section \ref*{#1}}}
\newcommand{\markAuthor}[1]{#1}

\begin{document}
\title{\vspace{-0.5em} {\scriptsize This work has been submitted to the IEEE for possible publication.}\\
	Considerations on the Evaluation of Biometric Quality Assessment Algorithms}
\author{
Torsten Schlett,
Christian Rathgeb,
Juan Tapia,
and
Christoph Busch%
\thanks{T. Schlett, C. Rathgeb, J. Tapia and C. Busch are with the  da/sec - Biometrics and Security Research Group, Hochschule Darmstadt, Germany, \{torsten.schlett, christian.rathgeb, juan.tapia-farias, christoph.busch\}@h-da.de}%
}

\maketitle

\begin{abstract}
Quality assessment algorithms can be used to estimate the utility of a biometric sample for the purpose of biometric recognition.
``Error versus Discard Characteristic'' (EDC) plots,
and ``partial Area Under Curve'' (pAUC) values of curves therein,
are generally used by researchers to evaluate the predictive performance of such quality assessment algorithms.
An EDC curve depends on 
an error type such as the ``False Non Match Rate'' (FNMR),
a quality assessment algorithm,
a biometric recognition system,
a set of comparisons each corresponding to a biometric sample pair,
and a comparison score threshold corresponding to a starting error.
To compute an EDC curve, comparisons are progressively discarded based on the associated samples' lowest quality scores, and the error is computed for the remaining comparisons.
Additionally, a discard fraction limit or range must be selected to compute pAUC values, which can then be used to quantitatively rank quality assessment algorithms.

This paper discusses and analyses various details for this kind of quality assessment algorithm evaluation,
including general EDC properties,
interpretability improvements for pAUC values based on a hard lower error limit and a soft upper error limit,
the use of relative instead of discrete rankings,
stepwise \vs{} linear curve interpolation,
and normalisation of quality scores to a $[0, 100]$ integer range.
We also analyse the stability of quantitative quality assessment algorithm rankings based on pAUC values across varying pAUC discard fraction limits and starting errors,
concluding that higher pAUC discard fraction limits should be preferred.
The analyses are conducted both with synthetic data and with real face image and fingerprint quality assessment data,
with a focus on general modality-independent conclusions for EDC evaluations.
Various EDC alternatives are discussed as well.

Open source evaluation software is provided at \url{https://github.com/dasec/quality-assessment-evaluation}\footnote{Will be made available upon acceptance.}.
\end{abstract}

\begin{IEEEkeywords}
Biometrics,
biometric sample quality,
error versus discard characteristic.
\end{IEEEkeywords}

\IEEEpeerreviewmaketitle

\vspace{-2em}
\ifCLASSOPTIONcompsoc
\IEEEraisesectionheading{\section{Introduction}}
\else
\section{Introduction}
\fi

\IEEEPARstart{B}{iometric} %
recognition \cite{Vocabulary} performance depends on the quality of the used biometric samples \cite{Vocabulary} such as face or fingerprint images.
In this context ``quality'' refers specifically to biometric utility \cite{Vocabulary},
as opposed to other definitions such as factor-specific quality (\eg{} ``How blurry is one image?'') or subjective image quality (\eg{} ``How noticeable are lossy compression artefacts to a human?'').

There are quality assessment (QA) algorithms that assign one scalar quality score (QS) \cite{Vocabulary} to one given biometric sample,
with higher scores implying higher biometric utility.
Biometric comparisons \cite{Vocabulary} involve information stemming from two samples,
so a QA algorithm attempts to assess a single sample's quality in terms of its utility for an unknown number of comparisons against unknown other samples.
Thus a QS can be considered an estimate of the usefulness of the information that is extractable from the sample, irrespective of other samples.

This paper is mainly about the quantitative ranking of QA algorithms with respect to concrete biometric recognition systems via ``Error versus Discard Characteristic'' (EDC) plots,
specifically using the ``partial Area Under Curve'' (pAUC) values thereof.
The EDC has been established as the de facto standard for QA algorithm evaluations, and is currently being officially standardised in the next edition of ISO/IEC 29794-1\footnote{\url{https://www.iso.org/standard/79519.html}}.
Conceptually, an EDC curve for one QA algorithm shows how some concrete biometric recognition ``error'' changes as samples are progressively ``discarded'' via an increasing sample QS threshold.
The primary insights from the analyses in this paper do not rely on the properties of specific biometric modalities and should thus generalise to any modality, when the quality of individual biometric samples is assessed with respect to comparisons between these samples.
Experiments with real QA algorithms in this paper were conducted mainly in the context of face image quality assessment (FIQA) \cite{Schlett-FIQA-LiteratureSurvey-CSUR-2021}, with additional fingerprint quality assessment experiments serving as an example for another biometric modality.
Various experiments only show results for one modality to avoid unnecessary clutter and to stay within the page limit.

The paper is structured as follows:

\begin{itemize}
\item In the remainder of this introduction the EDC and pAUC value rankings are described further in \autoref{sec:edc} and \autoref{sec:pauc}, respectively.
\item \SectionRef{sec:real-setup} details the setup used for the experiments with real QA algorithm data.
\item \SectionRef{sec:interpolation} discusses why ``stepwise'' EDC curve interpolation should be preferred.
\item \SectionRef{sec:normalisation} examines the effect of quality score normalisation, in particular to a $[0, 100]$ integer range.
\item \SectionRef{sec:stability-real} analyses the stability of QA algorithm rankings based on FNM-EDC pAUC values across different pAUC discard limit and starting error configurations, using real data, and \autoref{sec:stability-synthetic} continues the analysis with synthetic data that enables the comparisons of actual rankings against expected rankings.
\item \SectionRef{sec:other-approaches} presents various alternatives to the common EDC evaluation approach, discusses why this paper focusses on the EDC, and points out parts of the paper that also apply to the other approaches.
\item \SectionRef{sec:conclusions} briefly summarises the primary conclusions from the paper.
\end{itemize}

\subsection{Error versus Discard Characteristic}
\label{sec:edc}

The concept of the ``Error versus Discard Characteristic'' (EDC) was introduced under the name ``Error versus Reject Characteristic'' (ERC) by \markAuthor{Grother and Tabassi} \cite{Grother-SampleQualityMetricERC-PAMI-2007},
and is also known as ``Error versus Reject Curve'' in the literature.
At the time of writing the next (third) edition of ISO/IEC 29794-1 is standardising this concept under the name EDC instead of ERC to avoid confusion with other meanings of the word ``reject'' \cite{Vocabulary} in the biometric context.

A concrete EDC instance with real data consists of the following parts:
\begin{itemize}
\item Specific ``error'':
Often the ``False Non Match Rate'' (FNMR) \cite{Vocabulary} is selected as the EDC's error \cite{Schlett-FIQA-LiteratureSurvey-CSUR-2021}\cite{Grother-SampleQualityMetricERC-PAMI-2007},
a configuration abbreviated as FNM-EDC,
which also is what we focus on in this paper.
\item Set of comparisons:
Each comparison corresponds to one pair of samples as input and one comparison score (CS) \cite{Vocabulary} as output.
The selected error determines which kind of comparisons are required to compute the EDC.
For the FNM-EDC only mated \cite{Vocabulary} comparisons are used, meaning that the corresponding samples in each pair must stem from the same biometric capture subject \cite{Vocabulary} and the same biometric instance \cite{Vocabulary} (e.g. from the same finger).
\item One biometric recognition system:
The recognition system computes the CSs.
One can technically involve multiple recognition systems in a single EDC plot,
but for clarity we will use exactly one recognition system in each EDC instance within this paper,
which also is common practise in the scientific literature.
Additionally be aware that all CSs used in this paper are similarity scores \cite{Vocabulary}, meaning that higher CS values imply higher similarity.

Besides the recognition system,
a CS threshold needs to be defined to reach one comparison decision (match or non-match) \cite{Vocabulary} for each CS.
For the FNM-EDC each non-match counts as one comparison error, the FNMR being the non-match count divided by the number of comparisons.
\item Set of QA algorithms:
Every QA algorithm computes one QS per sample.
\item One pairwise QS function:
This function computes a single pairwise QS from two sample QSs that correspond to one comparison.
The data points of an EDC curve for one QA algorithm are computed by progressively discarding comparisons based on the associated pairwise QS.

It is technically possible to use a different function for each QA algorithm,
or to use multiple different functions per QA algorithm to effectively create different QA algorithm variants.
But in this paper we use exactly one function,
namely the minimum of each pairs' sample QSs as the pairwise QS,
which is the de facto standard in the literature
since lower QSs are supposed to indicate lower biometric utility.

The selected function should typically reflect the discarding of samples, not comparisons,
so that the EDC simulates the operational discarding of samples before any comparisons could be conducted.
Taking the minimum of the sample QSs does reflect this use case, since a pair will be counted as discarded when one of the samples would be discarded in isolation.
As an alternative to using a pairwise QS function for the EDC computation,
one can discard samples (and the associated comparisons) by their QS,
which is functionally equivalent to this pairwise minimum QS formulation.

Choosing a function that mixes the sample QSs to get the pairwise QS would instead represent an operational scenario in which the pairwise QSs are used to discard already computable comparisons.
This would effectively constitute an augmentation of the biometric comparison by adding a certainty value.
In this scenario the usage of ``pairwise QA algorithms'' that utilise information from both samples as input could also be considered,
so this scenario would not necessarily have to be restricted to (sample) QA algorithms.
\end{itemize}

\begin{figure}
\centering
\begin{tabular}{c}
LFW \\
\includegraphics[width=0.97\linewidth]{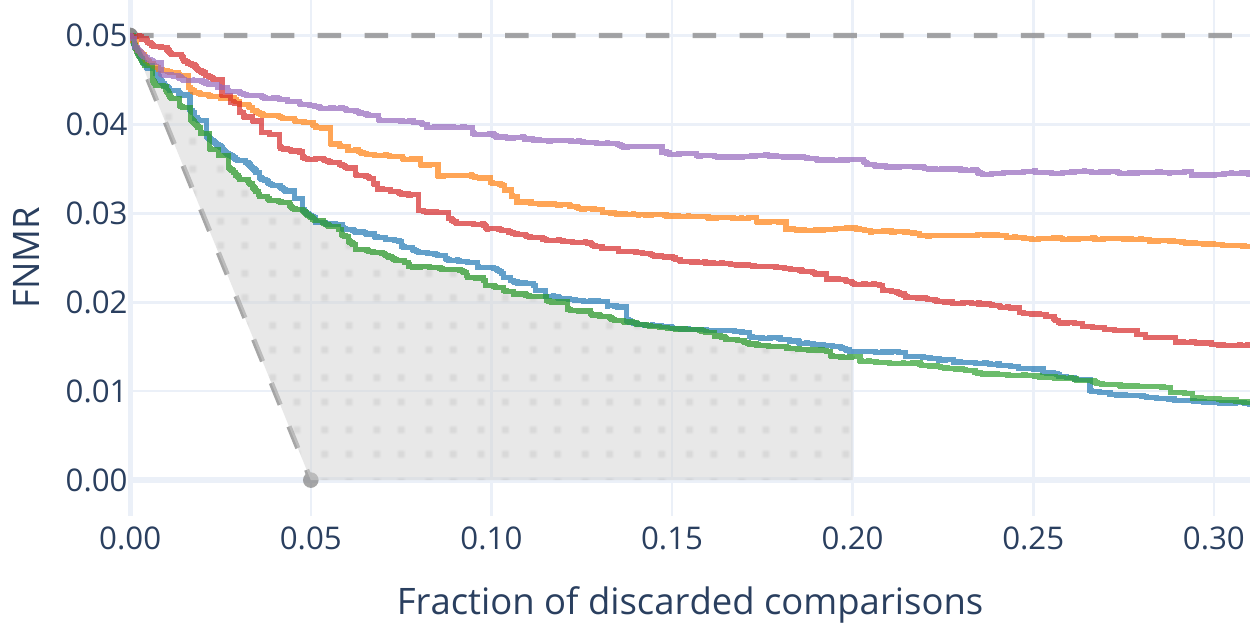} \\
TinyFace \\
\includegraphics[width=0.97\linewidth]{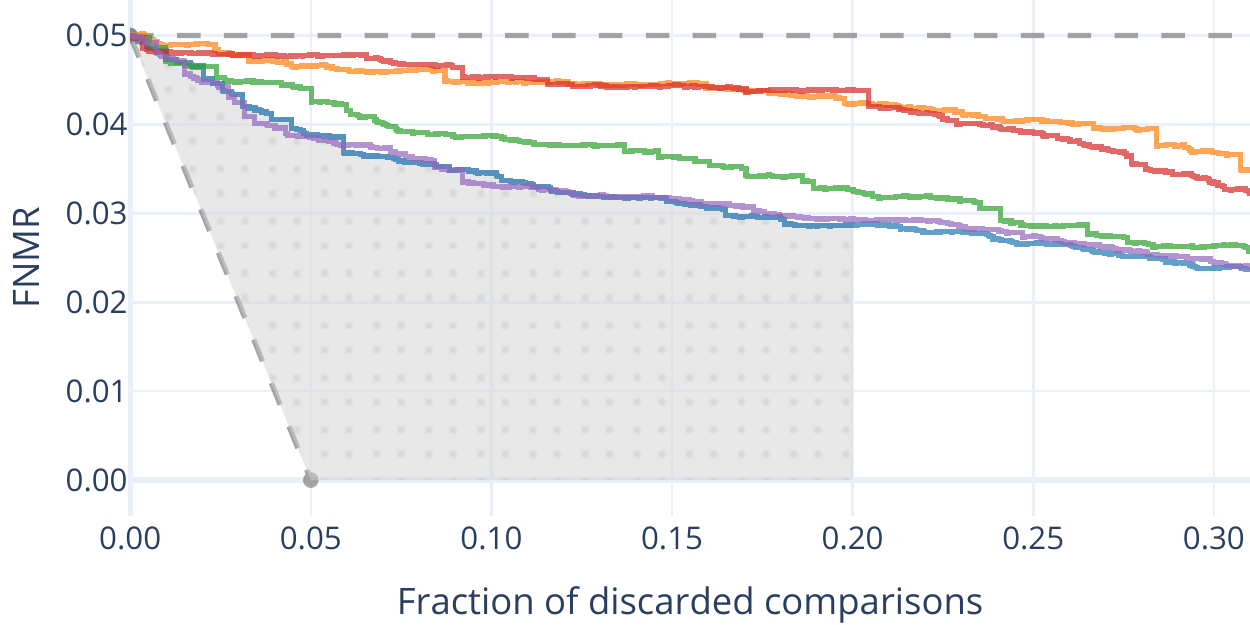} \\
\includegraphics[width=0.97\linewidth]{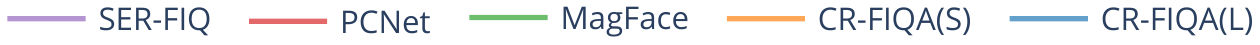} \\
\end{tabular}
\caption{\label{fig:edc-plot-example} FNM-EDC plot examples on different face image datasets with the same five QA algorithms, 0.05 starting error, and a shaded $[0,0.2]$ pAUC for the corresponding best curves (MagFace on LFW and CR-FIQA(L) on TinyFace), minus the ``area under theoretical best'' (below the lower grey dashed line). The used datasets and QA algorithms are described in \autoref{sec:real-setup}.}
\vspace{-2em}
\end{figure}

\begin{figure}
\centering
\begin{tabular}{c}
FVC2006-DB2\_A \\
\includegraphics[width=0.97\linewidth]{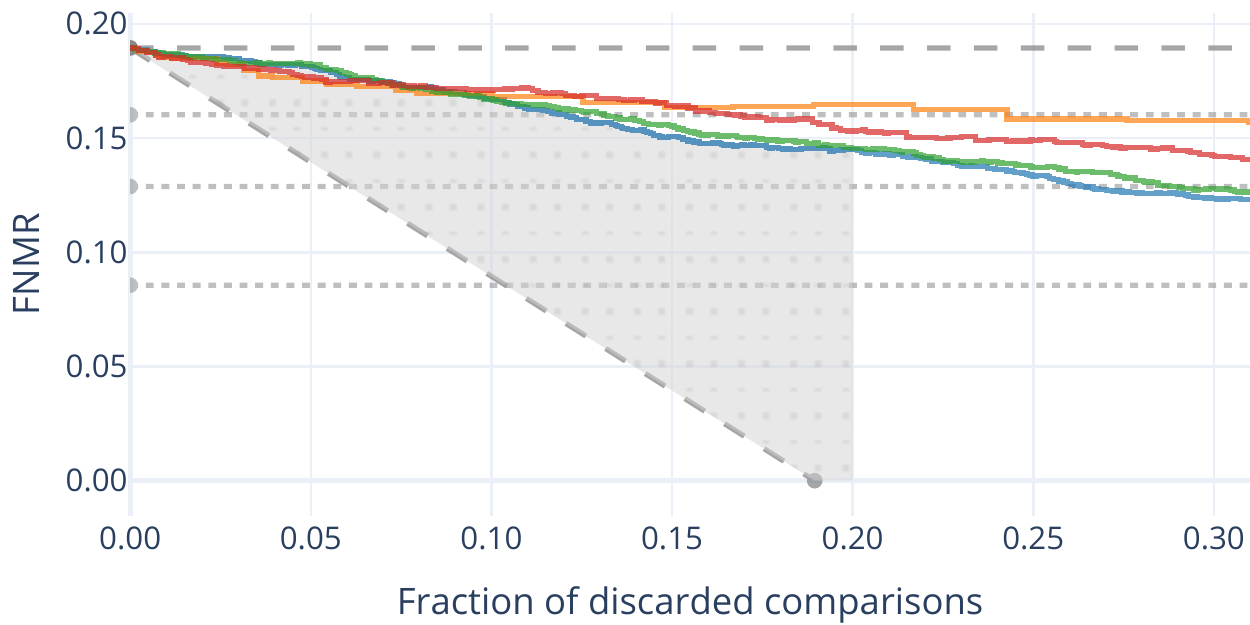} \\
\includegraphics[width=0.8\linewidth]{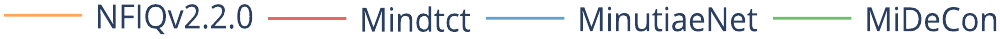} \\
\end{tabular}
\caption{\label{fig:edc-plot-example-fingerprint} An FNM-EDC plot example analogous to \autoref{fig:edc-plot-example}, but with the fingerprint setup described in \autoref{sec:real-setup}.
Dotted lines indicate other possible non-zero starting errors below the used one, which is approximately 0.1895.}
\vspace{-1em}
\end{figure}

\begin{figure}
\centering
\begin{tabular}{c}
LFW \\
\includegraphics[width=0.97\linewidth]{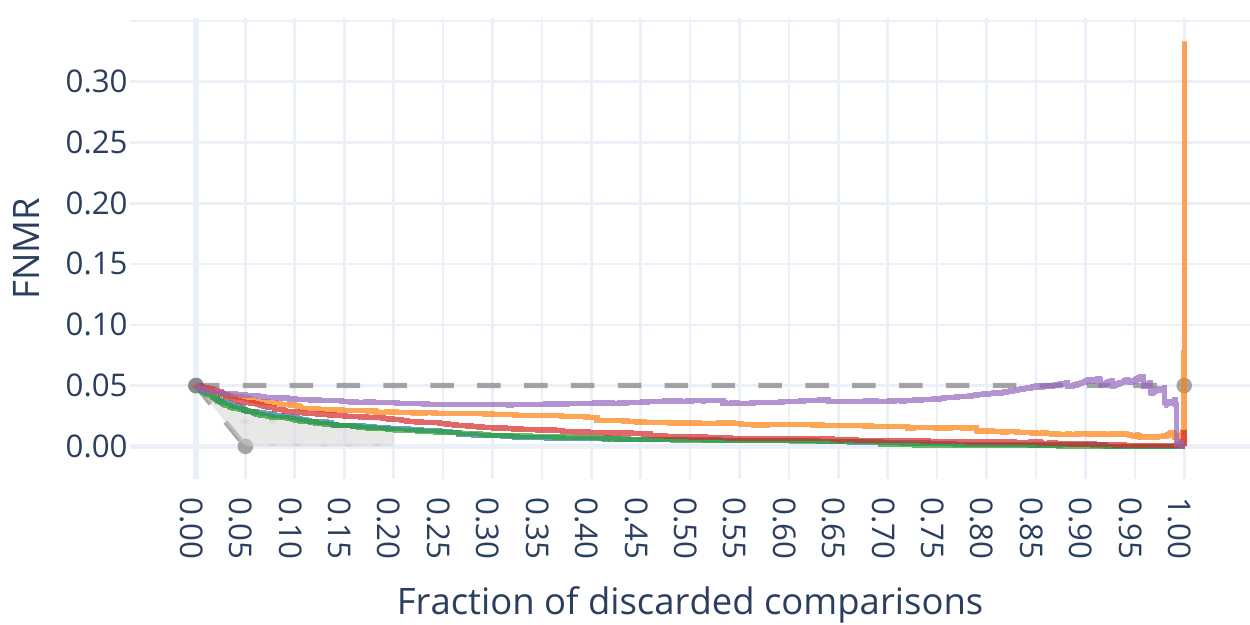} \\
\includegraphics[width=0.97\linewidth]{img/real-data/Real-FIQA-legend} \\
\end{tabular}
\caption{\label{fig:edc-plot-explosion} An example for ``exploding'' error values towards the 100\% discard fraction. This EDC plot shows the same curves as in \autoref{fig:edc-plot-example} for the LFW dataset, but for the full discard fraction range.}
\vspace{-1em}
\end{figure}

Each EDC curve corresponds to one QA algorithm.
Data points in the curve consist of the error and the discard fraction.
Conventionally the error is plotted on the Y-axis, increasing upwards, lower values being better,
while the discard fraction (of comparisons) is plotted increasing left to right on the X-axis.
Both the discard fraction and (usually) the error are inherently constrained to the range $[0,1]$.
See \autoref{fig:edc-plot-example} for an example.
To compute a curve the comparisons are progressively discarded in increasing order of their pairwise QSs.
Each discard step results in a curve data point with an implicit X-axis discard fraction,
for which the corresponding Y-axis error value is computed from the remaining comparisons.
To ensure that the curve is accurate, each discard step should only discard as many comparisons as strictly necessary to reach the next data point (\ie{} the set of comparisons with the next lowest pairwise QS),
as opposed to computing data points for \eg{} fixed QS threshold increments.
Note that the curve data point for discard fraction 0 is independent of the QA algorithms,
since it only depends on the comparison set and CS threshold,
meaning that there is one ``starting error'' per EDC plot in this paper.
See \autoref{sec:stability-real} for an analysis that involves varying starting errors.

\autoref{fig:edc-plot-example-fingerprint} shows an EDC plot example similar to \autoref{fig:edc-plot-example},
except using the fingerprint setup, which is further described in \autoref{sec:real-setup}.
While the used face image experiment data in \autoref{fig:edc-plot-example} allows for the selection of arbitrary starting errors with at least two digit precision, such as the used 0.05 value,
the fingerprint data in \autoref{fig:edc-plot-example-fingerprint} does not.
This fingerprint data serves as an example for a setup with too few different comparison scores to enable a flexible starting error selection.
The data is still sufficient to create an EDC plot as shown, but only using a relatively restricted number of starting errors.
In \autoref{fig:edc-plot-example-fingerprint} the possible non-zero starting errors below 0.20 are indicated by the dotted lines,
with the dashed line being the actually used starting error (approximately 0.1895).

If the full X-axis discard range is plotted, note that the Y-axis error values can - but do not have to - ``explode'' near the 100\% discard fraction,
meaning that the error values can be substantially larger than for the lower discard fractions.
\autoref{fig:edc-plot-explosion} shows a real example.
This can happen at these high discard fractions due to the lower number of remaining comparisons,
through which each individual comparison has a greater influence on the error percentage.
Thus even a low number of error comparisons can yield a high error percentage.

\subsection{Partial Area Under Curve}
\label{sec:pauc}

To quantitatively rank QA algorithms based on the EDC curves,
``partial Area Under Curve'' (pAUC) \cite{Olsen-FingerImageQuality-IETBiometrics-2016} values can be computed.
A pAUC value is the area under one curve for a chosen discard fraction range, \eg{} the $[0,0.2]$ range in \autoref{fig:edc-plot-example}.
Note that choosing higher discard fractions, \eg{} beyond 0.3, would not represent an operational scenario and should therefore be avoided.

Although pAUC values suffice to rank the QA algorithms relative to each other,
the magnitude of the differences might not necessarily be clearly interpretable,
since the raw pAUC values depend on the EDC starting error and the chosen discard fraction range.

\markAuthor{Olsen \etal{}} \cite{Olsen-FingerImageQuality-IETBiometrics-2016} proposed to subtract the ``area under theoretical best'' from the (p)AUC.
This refers to the area under the EDC curve for the theoretical best case where the decrease in the error equals the discard fraction\footnote{\markAuthor{Olsen \etal{}} \cite{Olsen-FingerImageQuality-IETBiometrics-2016} more specifically defined the area under theoretical best for FNM-EDC curves with the X-axis plotting the fraction of discarded samples instead of comparisons.},
\ie{} the area under the line defined by $max(0, StartingError-DiscardFraction)$,
which is also visible in \autoref{fig:edc-plot-example}.
Note that this is an approximation or lower limit of the theoretical best case, not the actual best case for the given comparison pairs.
The actual best case curve cannot be strictly monotonically decreasing,
since a real EDC curve can only change by discarding a non-fractional number of comparisons per data point,
which is further discussed and illustrated in \autoref{sec:interpolation}.

Subtracting the ``area under theoretical best'' from the pAUC values does not change the ranking of QA algorithms, since the same value is subtracted for each pAUC discard range configuration in which QA algorithms are ranked.
It can however serve as a straightforward\footnote{The ``area under theoretical best'' only depends on the EDC starting error.}
adjustment to make the pAUC values more easily interpretable,
since it removes the effect of the area that cannot possibly be improved.
The remaining pAUC, which is of interest, is the grey-coloured area in \autoref{fig:edc-plot-example}.

\begin{figure}
	\centering
	\includegraphics[width=\linewidth]{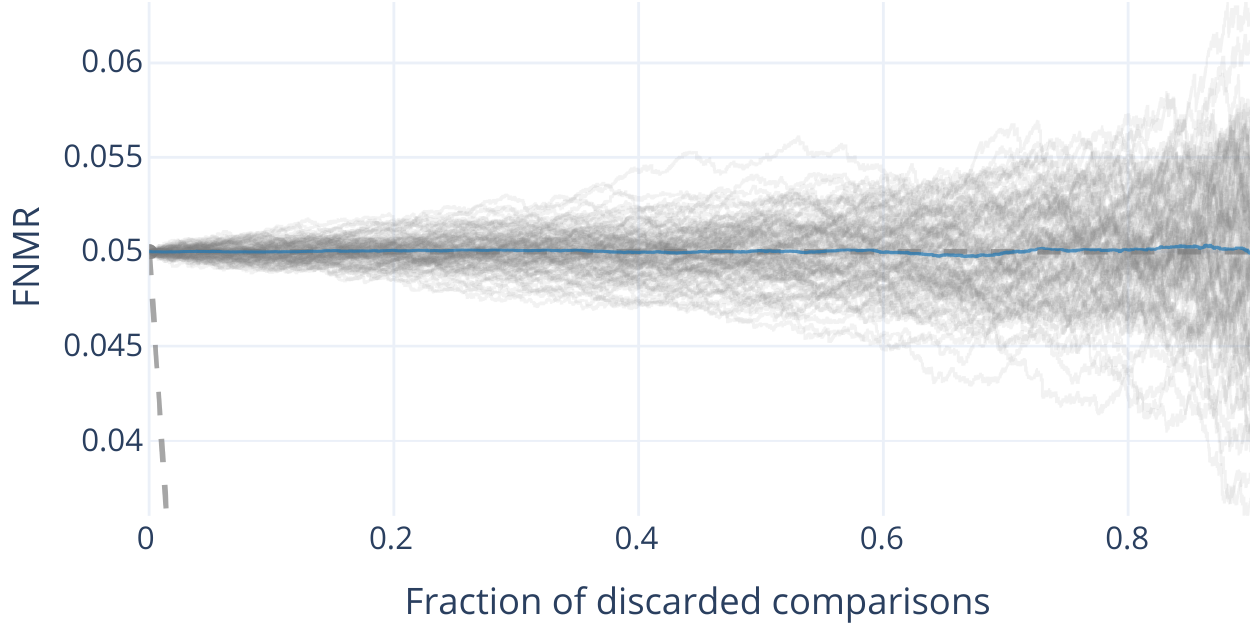}\\
	\includegraphics[width=0.37\linewidth]{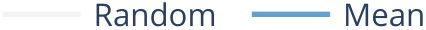}
	\caption{\label{fig:edc-random} The mean curve of many (here 100) EDC curves based on randomly generated QSs approximates the constant starting error (here 0.05), implying that QA algorithms resulting in curves distinctly above this constant are ineffective since they behave worse than (averaged) random QS curves.} %
	\vspace{-1em}
\end{figure}

In addition to this theoretical (hard) best case lower error bound line,
a theoretical (soft) worst case upper error bound line can be defined as well:
The constant EDC starting error line approximates the mean of infinite curves for random QSs,
as demonstrated in \autoref{fig:edc-random},
and a QA algorithm should of course preferably never increase the error above the starting error regardless of the discard fraction.
Therefore, the pAUC values for QA algorithms can optionally be made relative to the pAUC for this upper error bound line for the purposes of interpretability (\ie{} the ranking remains unaffected).

\begin{table}
\caption{\label{tab:qa-algorithm-ranking-example} Two QA algorithm ranking examples corresponding to the two EDC plots in \autoref{fig:edc-plot-example} for the $[0,0.2]$ pAUC range. The constant ``area under theoretical best'' value has been subtracted from the shown pAUC values, \ie{} here $(0.05^2)/2$ since the starting error is 0.05.}
\centering
\begin{tabular}{c}
LFW \\
\begin{tabular}{r|ccc}
\theader{QA algorithm} & \theader{pAUC value} & \theader{Discrete ranking} & \theader{Relative ranking} \\
\hline
               MagFace &              0.00362 &                          1 &                       0.00 \\
            CR-FIQA(L) &              0.00383 &                          2 &                       0.07 \\
                 PCNet &              0.00506 &                          3 &                       0.46 \\
            CR-FIQA(S) &              0.00572 &                          4 &                       0.68 \\
               SER-FIQ &              0.00672 &                          5 &                       1.00 \\
\end{tabular} \\
TinyFace \\
\begin{tabular}{r|ccc}
\theader{QA algorithm} & \theader{pAUC value} & \theader{Discrete ranking} & \theader{Relative ranking} \\
\hline
            CR-FIQA(L) &              0.00588 &                          1 &                       0.00 \\
               SER-FIQ &              0.00589 &                          2 &                       0.00 \\
               MagFace &              0.00666 &                          3 &                       0.38 \\
            CR-FIQA(S) &              0.00787 &                          4 &                       0.97 \\
                 PCNet &              0.00793 &                          5 &                       1.00 \\
\end{tabular}
\end{tabular}
\end{table}

These pAUC value interpretability adjustments are not necessary as long as the main concern of the evaluation is the performance of each QA algorithm relative to the other QA algorithms for a given EDC pAUC configuration.
In that case it can suffice to simply compute the QA algorithm ranking using the raw pAUC values,
see for example \autoref{tab:qa-algorithm-ranking-example}.
The ``relative ranking'' values are the min-max normalised pAUC values.
It shows how far an algorithm is considered from being the best (0) or worst (1) relative to the others,
in contrast to the ``discrete ranking''.
This ``relative ranking'' approach is used in the experiments of this paper.

Note that other evaluation types besides comparisons among multiple QA algorithms are possible, \eg{} a certification scheme that defines a certain pAUC value limit for a tested QA algorithm's EDC curve.

\section{Real data experiment setup}
\label{sec:real-setup}

This section describes the setup for the experiments that involve real instead of synthetic data.
As noted in the introduction,
there is a primary face image quality assessment (FIQA) setup
and a supporting fingerprint quality assessment setup.
The paper's general conclusions regarding QA algorithm evaluations using EDC curves should apply to scenarios with other biometric modalities as well,
since the underlying principles such as the FNMR, or the discarding of samples by a quality score threshold, are not specific to any modality.

\subsection{Used algorithms}

The real face image data experiments use one face detector for face image preprocessing, one face recognition system, and five (FI)QA algorithms:

\begin{itemize}
\item Face detector model: RetinaFace-R50 \cite{Deng-FaceDetection-RetinaFace-CVPR-2020}
\begin{itemize}
\item Images are excluded from the experiments when the face detection step fails.
\item The detected facial landmarks are used to preprocess the face images.
The same preprocessing approach used for ArcFace \cite{Deng-ArcFace-IEEE-CVPR-2019}
is also used for all FIQA models here, only the scaling differs since the models require different input resolutions.
Note that specialized preprocessing methods could be used for the individual models to possibly enhance their performance,
but that this work is about more general observations on the EDC, which should apply either way.
\end{itemize}
\item Face recognition feature extraction model: ArcFace-R100-MS1MV2 \cite{Deng-ArcFace-IEEE-CVPR-2019}.
\item FIQA algorithms:
\begin{itemize}
\item CR-FIQA(L) \cite{Boutros-FIQA-CRFIQA-arXiv-2022}: ResNet100 backbone trained on MS1MV2 \cite{Deng-ArcFace-IEEE-CVPR-2019}. $112\times 112$ input image size.
\item CR-FIQA(S) \cite{Boutros-FIQA-CRFIQA-arXiv-2022}: ResNet50 backbone trained on CASIA-WebFace \cite{Yi-LearningFaceRepresentationFromScratchCASIAWebFace-arXiv-2014}. $112\times 112$ input image size.
\item MagFace \cite{Meng-FRwithFQA-MagFace-CVPR-2021}: ResNet100 backbone trained on MS1MV2 \cite{Deng-ArcFace-IEEE-CVPR-2019}. $112\times 112$ input image size. %
\item PCNet \cite{Xie-FQA-PredictiveUncertaintyEstimation-BMVC-2020}: Trained on VGGFace2 \cite{Cao-VGGFace2Dataset-FGR-2018}. $224\times 224$ input image size.
\item SER-FIQ \cite{Terhorst-FQA-SERFIQ-CVPR-2020}: ``Same model'' variant using ArcFace. $112\times 112$ input image size.
\end{itemize}
\end{itemize}

The PCNet model was provided to us by one of the authors, the other models are publicly available.

The fingerprint setup is partially based on the MiDeCon paper by \markAuthor{Terhörst \etal{}} \cite{Terhorst-FingerQA-MiDeCon-IJCB-2021}:
\begin{itemize}
\item Fingerprint comparisons:
\begin{itemize}
\item Minutiae extractor: Mindtct (NBIS 5.0.0 \cite{NBIS-User-Guide}).
\item Minutiae comparator: Bozorth3 (NBIS 5.0.0 \cite{NBIS-User-Guide}), using the top 20 (or less) minutiae.
\end{itemize}
\item Fingerprint QA algorithms:
\begin{itemize}
\item NFIQ v2.2.0 \cite{NIST-NFIQ2-FingerprintImageQuality-2021}.
\item Mindtct (NBIS 5.0.0 \cite{NBIS-User-Guide}).
\item MinutiaeNet \cite{Nguyen-MinutiaeNet-ICB-2018}.
\item MiDeCon \cite{Terhorst-FingerQA-MiDeCon-IJCB-2021}.
\end{itemize}
\end{itemize}
Mindtct, MinutiaeNet, and MiDeCon detect minutiae with confidence values.
The mean of the top 20 (or less) minutiae detection confidence values is used as the sample QS.

\subsection{Used datasets}

Two face image datasets are used:

\begin{itemize}
\item LFW (Labeled Faces in the Wild) \cite{LFWTech}
\begin{itemize}
\item Type: Web-scraped (varying quality).
\item Image width $\times$ height: $250\times 250$
\item Mean face region width $\times$ height: $94.98\times 129.63$
\item Excluded image file duplicates: 2
\item Images subsequently excluded due to failed face detection: 0
\item Images remaining: 13,231
\item Remaining subjects with only one image (implicitly excluded from mated pair set): 4,067
\item Subjects in mated comparisons: 1,680
\item Images in mated comparisons: 9,164
\item Mated comparisons used: 242,257
\end{itemize}
\item TinyFace (subsets Testing\_Set/Gallery\_Match and Testing\_Set/Probe) \cite{Cheng-TinyFace-LowResolutionFaceRecognition-ACCV-2018}
\begin{itemize}
\item Type: Web-scraped (varying quality).
\item Mean image width $\times$ height: $30.64\times 31.74$
\item Mean face region width $\times$ height: $29.22\times 31.50$
\item Excluded image file duplicates: 184
\item Images subsequently excluded due to failed face detection: 132
\item Images remaining: 7,855
\item Remaining subjects with only one image (implicitly excluded from mated pair set): 131
\item Subjects in mated comparisons: 2,434
\item Images in mated comparisons: 7,724
\item Mated comparisons used: 19,478
\end{itemize}
\end{itemize}

The number of used mated comparison pairs is the number of all possible mated pairs for the images which were not excluded.
Lists of the excluded image file duplicates are provided in supplemental material\footnote{\url{https://github.com/dasec/dataset-duplicates}}, due to the larger number of duplicates found in TinyFace.

Some of the example plots in \autoref{sec:other-approaches} incorporate non-mated comparisons besides the mated comparisons.
A number of non-mated comparisons equal to the number of mated comparisons is randomly selected for these cases.

For the fingerprint setup, the ``DB2\_A'' subset of FVC 2006 \cite{FVC-EvaluationReport-2006} is used:
\begin{itemize}
\item Type: Optical sensor, 596dpi
\item Images: 1,680
\item Subjects: 140 (12 images each)
\item Mated comparisons: 9,240 (all mated pairs are used)
\end{itemize}

\section{Curve interpolation}
\label{sec:interpolation}

\begin{figure}
\begin{tabular}{c}
Stepwise interpolation \\
\includegraphics[width=\linewidth]{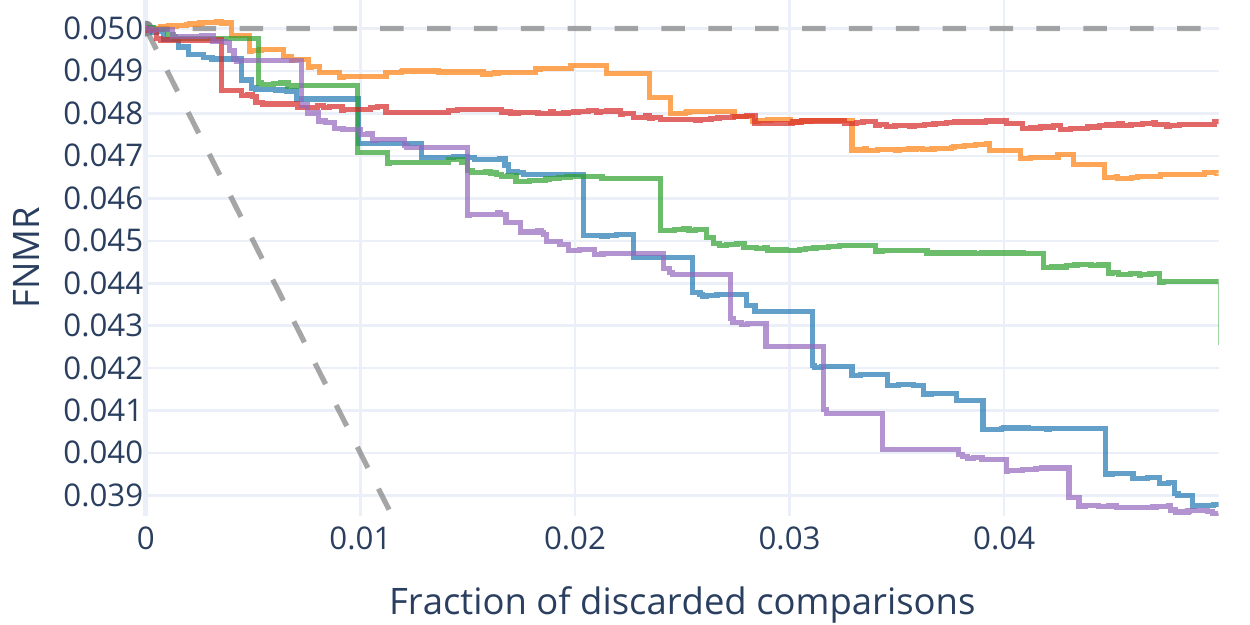} \\
Linear interpolation \\
\includegraphics[width=\linewidth]{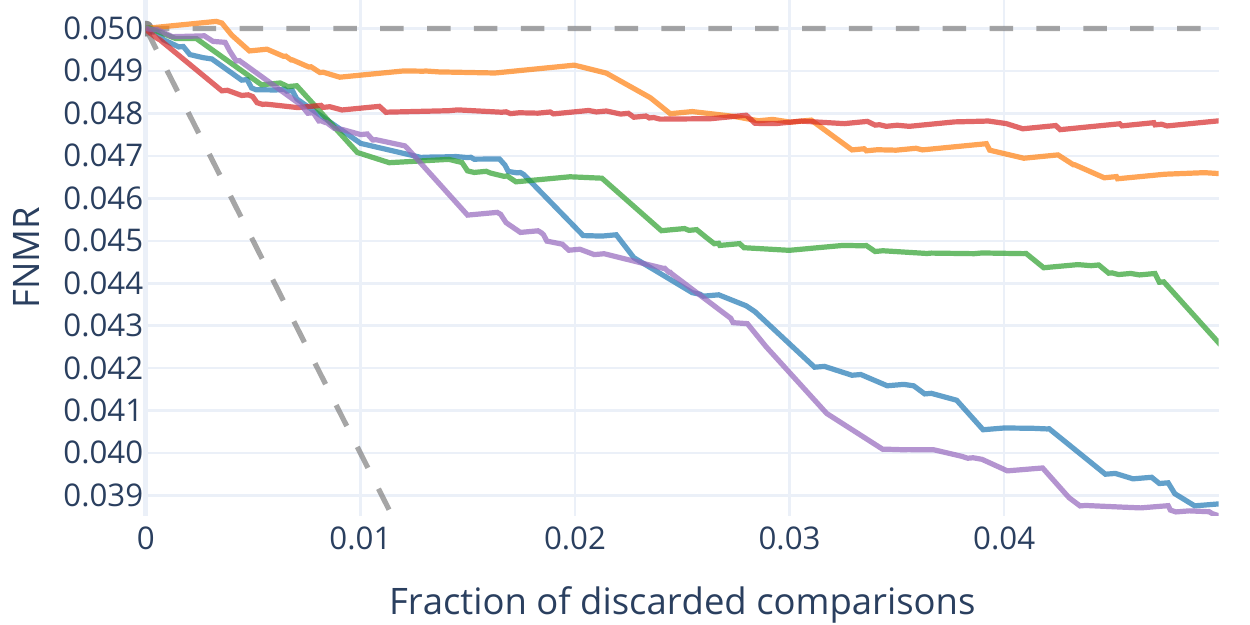} \\
\includegraphics[width=0.96\linewidth]{img/real-data/Real-FIQA-legend} \\
\end{tabular}
\caption{\label{fig:edc-interpolation}Stepwise \vs{} linear EDC curve interpolation example with real data.} %
\vspace{-2em}
\end{figure}

\begin{figure}
\begin{tabular}{c}
a) 10 steps at starting error 0.50 \\
\includegraphics[width=0.97\linewidth]{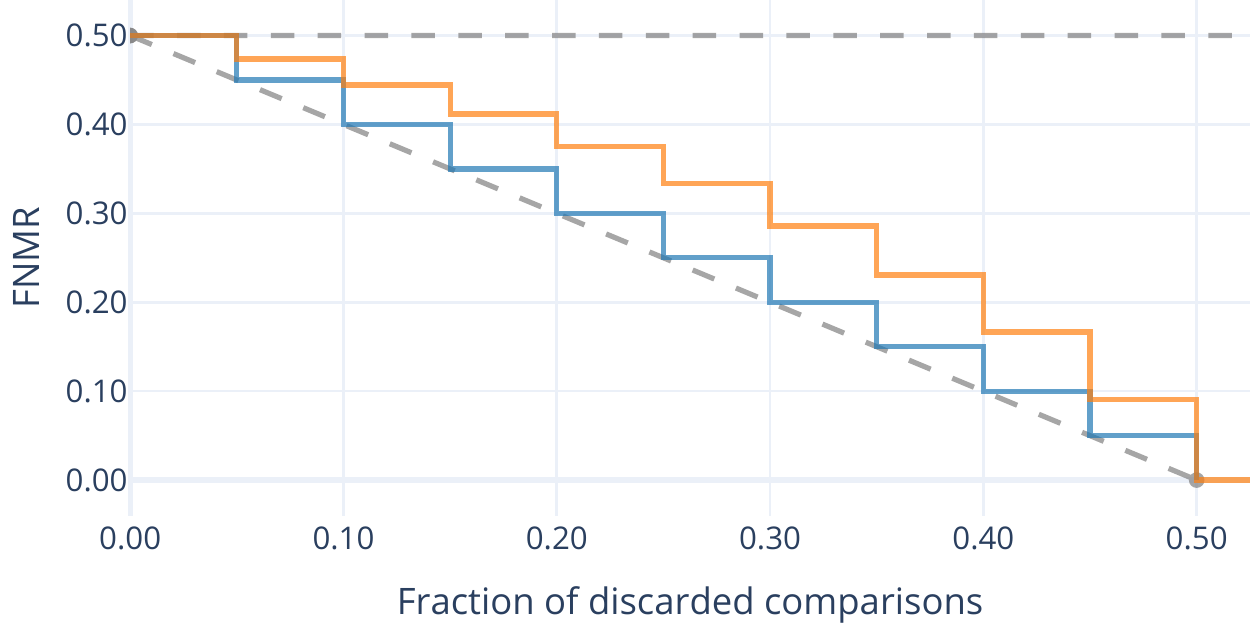} \\
b) 50 steps at starting error 0.50 \\
\includegraphics[width=0.97\linewidth]{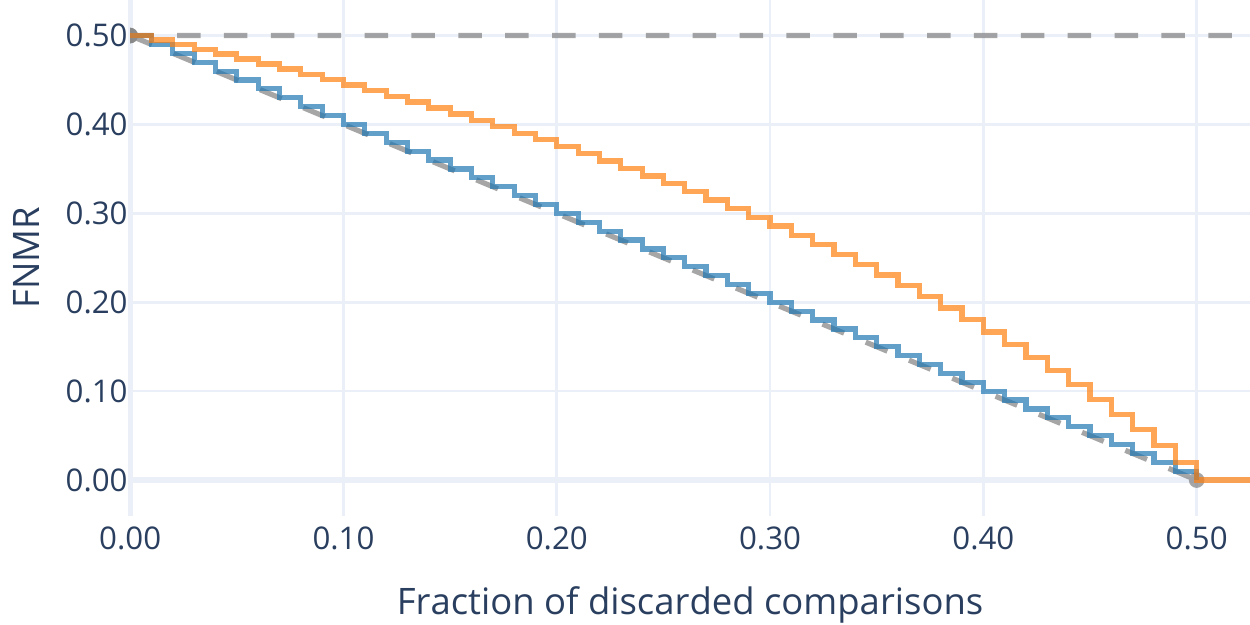} \\
c) 50 steps at starting error 0.05 \\
\includegraphics[width=0.97\linewidth]{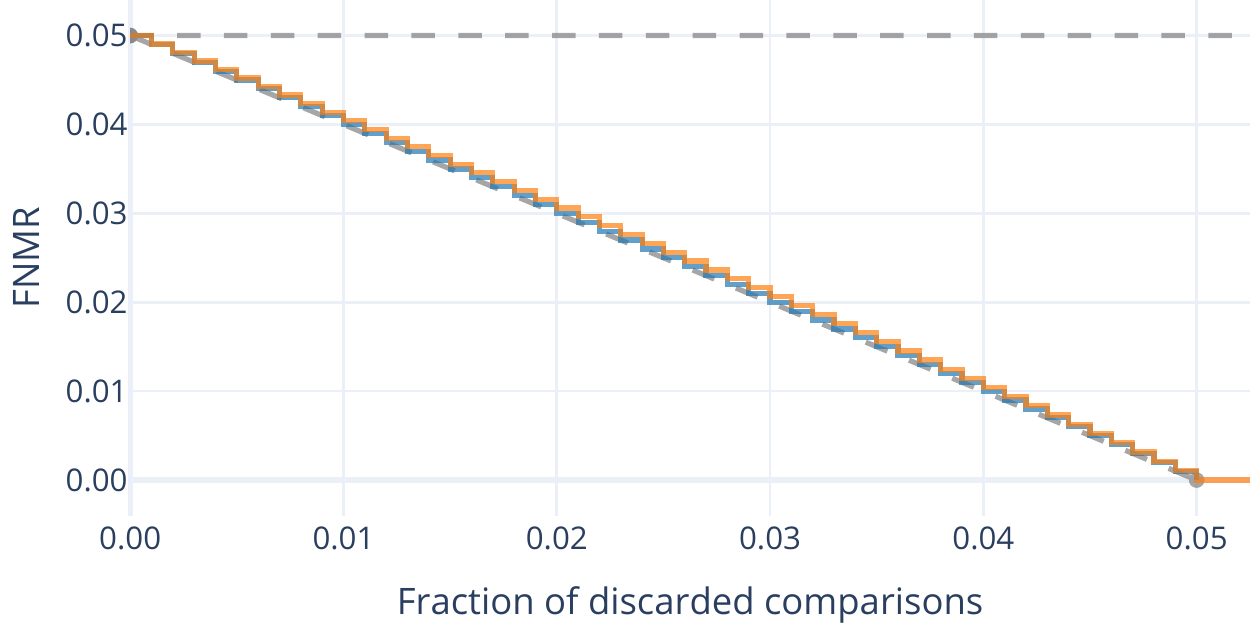} \\
\includegraphics[width=0.65\linewidth]{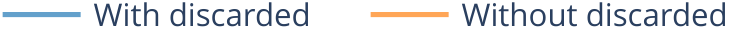} \\
\end{tabular}
\caption{\label{fig:edc-theoretical-best-steps} Examples for best possible EDC curves (coloured) in comparison to the dashed lower grey ``theoretical best'' line.}
\vspace{-2em}
\end{figure}

As described in \autoref{sec:edc},
every data point in an EDC curve corresponds to a discrete number of discarded comparisons.
This means that the fraction of discarded comparisons that is plotted on the X-axis technically is the fractional upper limit of discarded comparisons,
which maps to a discrete number of actually discarded comparisons.
To reflect this property,
we recommend using ``stepwise'' curve interpolation for EDC plots,
meaning that the curve's error (Y-axis) value only changes at each concrete data point.
Using linear interpolation instead may be misleading, which is demonstrated by the interpolation-dependent curve intersection points in \autoref{fig:edc-interpolation}.
Note that the curve interpolation choice also affects the pAUC computation, and thus QA algorithm rankings based on the pAUC values.

\autoref{fig:edc-theoretical-best-steps} further shows how a best possible EDC curve with stepwise interpolation in comparison to the ``theoretical best'' line may look.
For illustrative purposes the \autoref{fig:edc-theoretical-best-steps} a) plot shows curves with only ten equally sized ``steps'', corresponding to an equal number of comparisons for ten discard steps.
The Y-axis error values (FNMR) for the blue curves are identical to their corresponding X-axis discard fraction value,
which means that the error is computed by dividing the number of error cases (CSs below the threshold for FNMR) by the constant total number of comparisons,
thus as labelled the error is computed ``With discarded'' comparisons included. 
For the orange curves the error is instead computed by dividing the number of error cases by the number of remaining comparisons, \ie{} ``Without discarded'' comparisons.
The latter ``Without discarded'' error computation is used within this paper and should generally be preferred, since it corresponds to a scenario in which samples are discarded before they would be involved in comparisons.
In contrast, the ``With discarded'' computation would be unable to show increasing errors due to the discarding of non-error (true positive) comparisons, the denominator being a constant and the numerator being the error count that cannot increase.
As visible in the exaggerated case in the \autoref{fig:edc-theoretical-best-steps} a) plot,
the Y-axis error values for a best case curve ``Without discarded'' can technically deviate more substantially from the lower dashed theoretical best line than the best case curve ``With discarded'', despite the same X-axis discard steps.
This is because the denominator used to compute the FNMR, \ie{} the number of remaining comparisons, decreases with increasing discard fraction.
An increased number of discard steps cannot eliminate the difference of the ``Without discarded'' curve to the theoretical best line since this difference depends on the starting error, as the \autoref{fig:edc-theoretical-best-steps} b) plot exemplifies.
A lower starting error can however reduce the difference, as the \autoref{fig:edc-theoretical-best-steps} c) plot shows.
This detail of the best possible real (``Without discarded'') curve behaviour may thus not be important to consider in practice when lower starting errors and larger numbers of comparisons are used,
since these allow for a better approximation of the theoretical best line.
But note that real EDC curves as shown \eg{} in \autoref{fig:edc-interpolation} may also involve differently sized discard steps, \ie{} the discarding of different numbers of comparisons, which can happen in different orders depending on the (pairwise) QSs.

\section{Quality score normalisation}
\label{sec:normalisation}

The ``raw'' QSs produced by different QA algorithms can be numbers that lie in various ranges with different granularity.
For example, the QSs produced by the real FIQA algorithms described in \autoref{sec:real-setup} are floating point numbers in various ranges.
The presence of different QS ranges per QA algorithm is unproblematic for the computation of EDC curves in a plot, provided the QSs from different QA algorithms aren't mixed for a curve, since the EDC curves depend only on the (discard) order of the (pairwise) QSs relative to each other.
Different QA algorithm output ranges and QS distributions are however relevant if the QSs from different algorithms should be made similarly interpretable (\eg{} to apply the same QS threshold to discard samples across multiple QA algorithms), or to fuse QSs from different QA algorithms \cite{Schlett-FIQA-Fusion-BIOSIG-2022}, or because the ``raw'' QS output isn't usable for a certain data format.

A concrete instance of the latter scenario can be found in ISO/IEC 29794-1:2016 \cite{ISO-IEC-29794-1-QualityFramework-160915},
which prescribes a $[0,100]$ integer range for QSs as a mandatory requirement of the standardised data interchange format \cite{ISO-IEC-39794-1-G3-Framework-191223}\cite{ISO-IEC-39794-5-G3-FaceImage-191015}.
QSs can be normalised to such integer ranges, and this can also be done before an EDC evaluation is carried out.
This section examines the effect of this normalisation on the EDC curves across different normalisation configurations using the real \autoref{sec:real-setup} FIQA data, since this data comprises two datasets with distinct QS distributions (as will be illustrated).

Note that some QA algorithms do not need any separate normalisation, as is the case for NFIQ 2 \cite{NIST-NFIQ2-FingerprintImageQuality-2021} in the fingerprint setup.
NFIQ 2 outputs QSs in the $[0,100]$ integer range, and is formally recognized as a reference implementation of the normative measures presented in ISO/IEC 29794-4 \cite{ISO-IEC-29794-4-FingerQuality-170601}\footnote{See \url{https://github.com/usnistgov/NFIQ2}.}.

To normalise the ``raw'' QSs to $[0,100]$ integers (\ie{} 101 bins), 100 QS boundaries are calibrated based on a certain calibration function and a set of raw QSs.
The set of raw QSs is used as input for the calibration function.
Three different raw calibration QS set variants are part of the analysis:
\begin{itemize}
\item ``Same'': The same raw QSs used for the unnormalised EDC curves are used for calibration. \Ie{} the raw QSs for one QA algorithm on one dataset are used both to compute the EDC curve without normalisation, and to create the normalised variant thereof. This can therefore be considered as a best-case scenario in which the calibration data is equivalent to the evaluation data.
\item ``Other'': Raw QSs from the ``other'' of the two datasets (LFW and TinyFace) are used for calibration. \Ie{} to compute the EDC curve on LFW for one QA algorithm with normalisation, raw QSs for the same QA algorithm from TinyFace are used for the calibration (and vice versa). The QS distributions between these two datasets can differ substantially, as visible in the following results, so this represents a calibration scenario in which the calibration data is suboptimal.
\item ``Combined'': Here the raw QS sets from ``Same'' and ``Other'' (\ie{} from both datasets) are merged for calibration.
This represents a scenario wherein a broader coverage of calibration data across datasets is used, and where the optimal set of QSs happens to be included in the calibration data among other QSs.
\end{itemize}

\begin{figure}
\centering
\begin{tabular}{c}
	MinMax \\
	\includegraphics[width=0.97\linewidth]{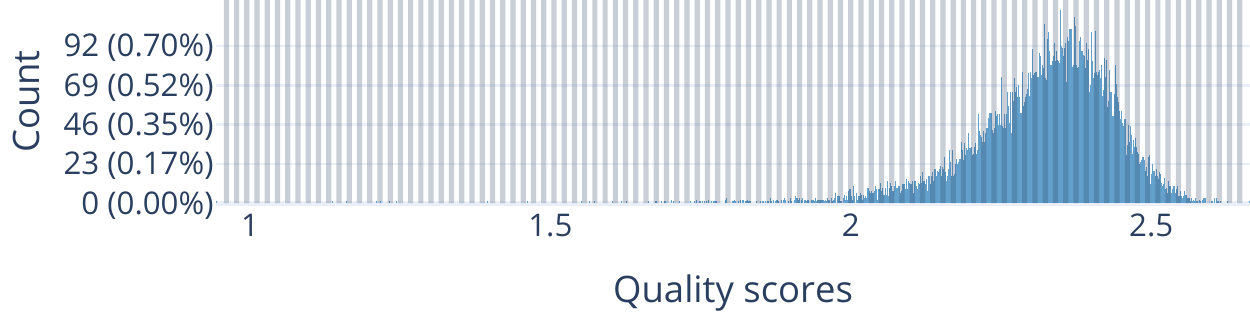} \\
	Proportional \\
	\includegraphics[width=0.97\linewidth]{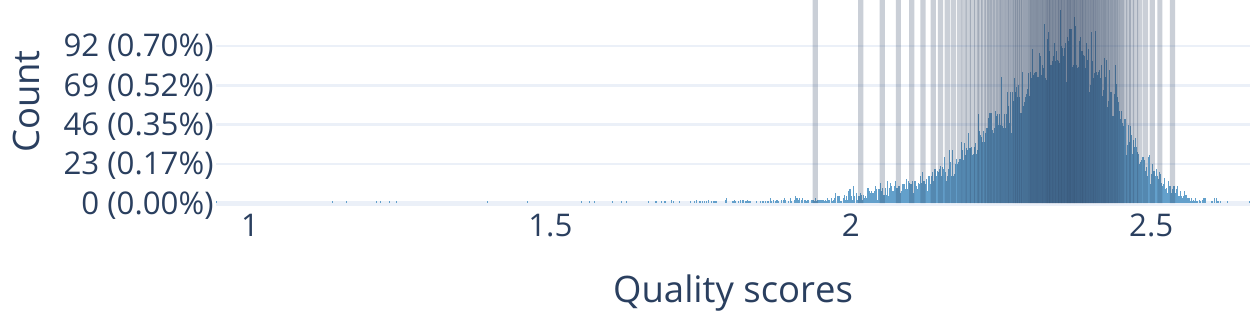} \\
\end{tabular}
\caption{\label{fig:normalisation-dist-same-minmax-vs-proportional} Comparison between ``MinMax'' and ``Proportional'' calibration of the $[0, 100]$ QS normalisation boundaries, based on the same QS scores from CR-FIQA(L) on the LFW dataset for example. The Y-axis shows QS histogram bin counts, with the share of the total QS count in parentheses.}
\vspace{-1em}
\end{figure}

\begin{figure}
\centering
\begin{tabular}{c}
	LFW \\
	\includegraphics[width=0.97\linewidth]{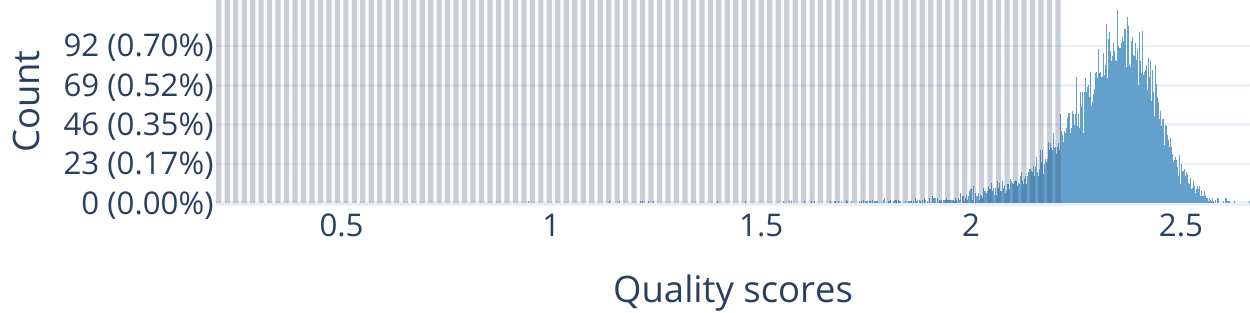} \\
	TinyFace \\
	\includegraphics[width=0.97\linewidth]{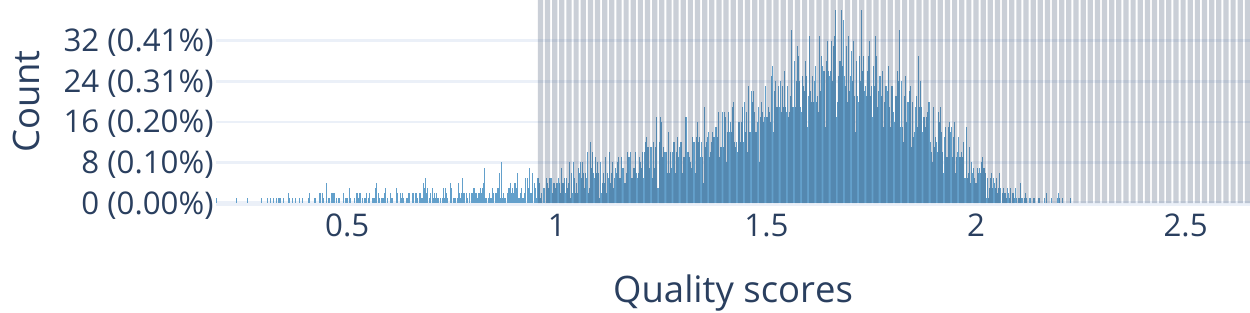} \\
\end{tabular}
\caption{\label{fig:normalisation-dist-other-lfw-vs-tinyface} $[0, 100]$ QS normalisation boundaries obtained by ``MinMax'' calibration on QSs for the other dataset (LFW for TinyFace, TinyFace for LFW), using CR-FIQA(L) for example. The Y-axis shows QS histogram bin counts, with the share of the total QS count in parentheses.}
\vspace{-1.5em}
\end{figure}

The shown analysis further includes two different calibration functions:
\begin{itemize}
\item ``MinMax'': The range between the minimum and the maximum of the raw QSs used for calibration is equally subdivided. All other calibration QSs simply have no effect on the calibration.
\item ``Proportional'': The normalisation QS boundaries are fitted to the distribution of the raw calibration QSs so that (approximately) the same number of calibration QSs is mapped to each of the 101 normalised QSs. \Ie{} the density of the calibration QSs is reflected in the resulting normalisation QS boundaries.
\end{itemize}

\begin{figure}
	\centering
	\begin{tabular}{c}
		Same \\
		\includegraphics[width=\linewidth]{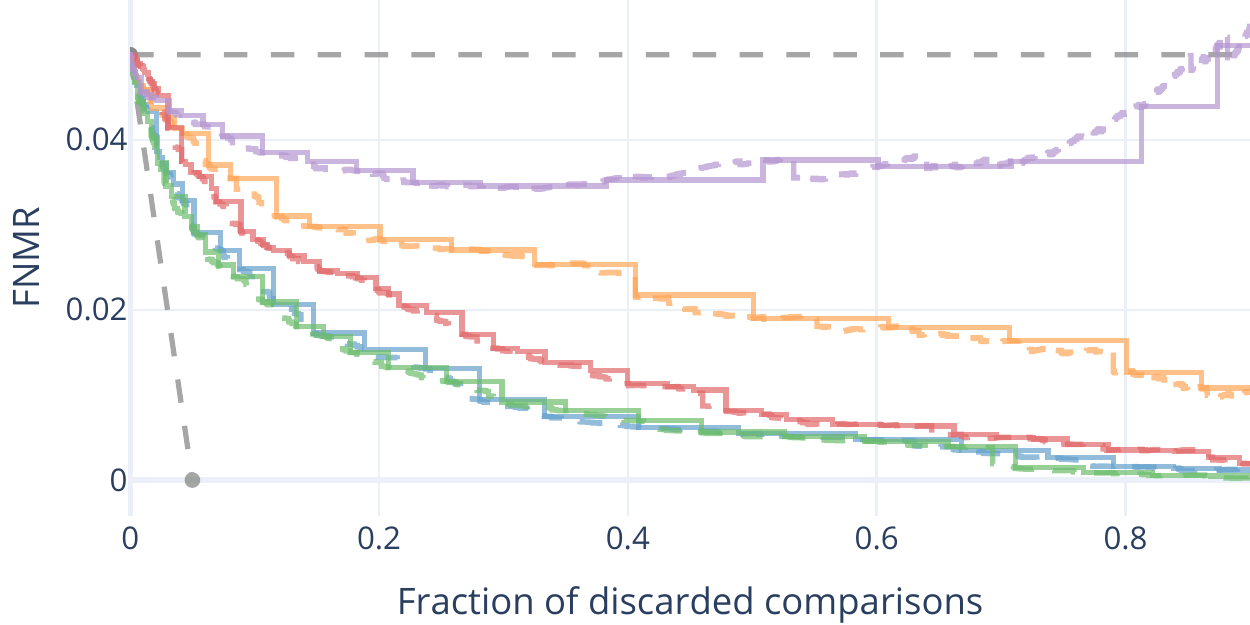} \\
		Other \\
		\includegraphics[width=\linewidth]{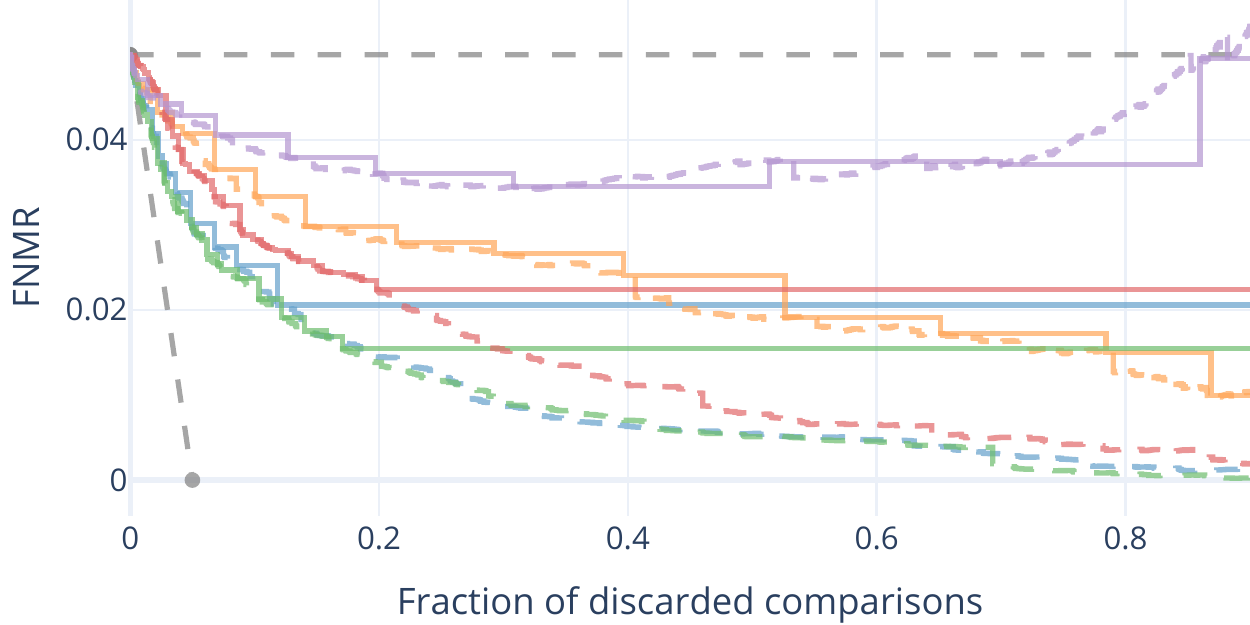} \\
		\includegraphics[width=0.87\linewidth]{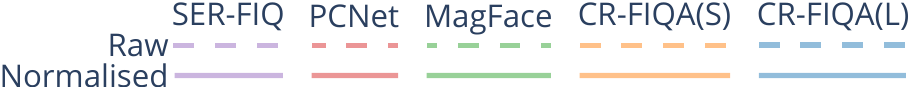} \\
	\end{tabular}
	\caption{\label{fig:normalisation-edc-lfw-same-vs-other} Two EDC plot examples on LFW with curves for raw QSs and QSs normalised based on ``MinMax'' calibration, one using calibration QSs from the same dataset (LFW) and one from another dataset (TinyFace).}
	\vspace{-1.5em}
\end{figure}

\begin{table*}
	\centering
	\caption{\label{tab:normalisation} The divergence of EDC curves based on normalised QSs from the EDC curves based on the corresponding raw QSs, expressed as the area between the curves divided by the raw QS curve pAUC value, in percent (\ie{} always zero or positive, lower is better).}
	\scriptsize
	\begin{tabular}{cccc|c|ccccc}
		pAUC discard range & Calibration data & Calibration function & EDC data &  Mean & CR-FIQA(L) & CR-FIQA(S) & MagFace & PCNet & SER-FIQ \\
		\hline
		    $[0.00, 0.20]$ &             Same &               MinMax &      LFW &  2.78\% &       4.14\% &       2.97\% &    3.53\% &  1.71\% &    1.55\% \\
		    $[0.00, 0.20]$ &             Same &               MinMax & TinyFace &  0.79\% &       0.70\% &       0.82\% &    1.29\% &  0.33\% &    0.83\% \\
		    $[0.00, 0.20]$ &             Same &         Proportional &      LFW &  2.13\% &       2.98\% &       2.47\% &    2.64\% &  1.98\% &    0.60\% \\
		    $[0.00, 0.20]$ &             Same &         Proportional & TinyFace &  2.75\% &       4.79\% &       1.06\% &    2.36\% &  1.07\% &    4.48\% \\
		\hline
		    $[0.00, 0.20]$ &            Other &               MinMax &      LFW &  3.74\% &       8.44\% &       3.76\% &    2.53\% &  1.15\% &    2.80\% \\
		    $[0.00, 0.20]$ &            Other &               MinMax & TinyFace & 11.06\% &      28.99\% &       9.72\% &    1.75\% &  7.74\% &    7.09\% \\
		    $[0.00, 0.20]$ &            Other &         Proportional &      LFW & 24.56\% &      55.13\% &       8.99\% &   35.95\% & 21.84\% &    0.86\% \\
		    $[0.00, 0.20]$ &            Other &         Proportional & TinyFace & 25.09\% &      40.22\% &       9.72\% &   26.40\% &  8.97\% &   40.14\% \\
		\hline
		    $[0.00, 0.20]$ &         Combined &               MinMax &      LFW &  4.17\% &       7.20\% &       4.57\% &    3.88\% &  2.17\% &    3.04\% \\
		    $[0.00, 0.20]$ &         Combined &               MinMax & TinyFace &  0.98\% &       0.94\% &       0.70\% &    1.97\% &  0.42\% &    0.88\% \\
		    $[0.00, 0.20]$ &         Combined &         Proportional &      LFW &  2.49\% &       3.61\% &       2.85\% &    3.22\% &  2.20\% &    0.56\% \\
		    $[0.00, 0.20]$ &         Combined &         Proportional & TinyFace &  7.94\% &      14.10\% &       2.62\% &    6.47\% &  2.76\% &   13.74\% \\
		\hline
		    $[0.00, 0.30]$ &             Same &               MinMax &      LFW &  3.29\% &       5.61\% &       2.64\% &    4.44\% &  2.25\% &    1.52\% \\
		    $[0.00, 0.30]$ &             Same &               MinMax & TinyFace &  1.27\% &       0.92\% &       1.21\% &    1.86\% &  1.32\% &    1.04\% \\
		    $[0.00, 0.30]$ &             Same &         Proportional &      LFW &  1.98\% &       2.82\% &       1.97\% &    2.57\% &  1.98\% &    0.56\% \\
		    $[0.00, 0.30]$ &             Same &         Proportional & TinyFace &  2.44\% &       3.97\% &       1.09\% &    2.26\% &  1.23\% &    3.65\% \\
		\hline
		    $[0.00, 0.30]$ &            Other &               MinMax &      LFW &  8.27\% &      20.82\% &       3.53\% &    8.36\% &  5.65\% &    3.02\% \\
		    $[0.00, 0.30]$ &            Other &               MinMax & TinyFace & 10.04\% &      21.43\% &      14.15\% &    2.28\% &  6.94\% &    5.38\% \\
		    $[0.00, 0.30]$ &            Other &         Proportional &      LFW & 40.22\% &      87.53\% &      12.53\% &   61.55\% & 38.65\% &    0.84\% \\
		    $[0.00, 0.30]$ &            Other &         Proportional & TinyFace & 34.57\% &      53.19\% &      14.15\% &   38.47\% & 15.04\% &   52.00\% \\
		\hline
		    $[0.00, 0.30]$ &         Combined &               MinMax &      LFW &  4.56\% &       8.72\% &       3.86\% &    4.63\% &  2.82\% &    2.78\% \\
		    $[0.00, 0.30]$ &         Combined &               MinMax & TinyFace &  1.75\% &       1.05\% &       1.10\% &    3.39\% &  2.17\% &    1.05\% \\
		    $[0.00, 0.30]$ &         Combined &         Proportional &      LFW &  2.47\% &       3.74\% &       2.27\% &    3.31\% &  2.45\% &    0.56\% \\
		    $[0.00, 0.30]$ &         Combined &         Proportional & TinyFace &  6.93\% &      11.42\% &       2.69\% &    5.95\% &  3.49\% &   11.07\% \\
	\end{tabular}
\end{table*}

\autoref{fig:normalisation-dist-same-minmax-vs-proportional} illustrates the difference between the ``MinMax'' and ``Proportional'' calibration, for the QA algorithm CR-FIQA(L) on the LFW dataset (\ie{} using the ``Same'' calibration data variant).
The plots show that the ``Proportional'' calibration neglects detail for a relatively large range of lower QSs,
whereas the ``MinMax'' calibration doesn't.
Be aware that these plots include the entire QS distribution, meaning that the X-axis shows the entire range from the QS minimum to the QS maximum, but the minimum data point is not easily visible due to a low Y-axis value.

\autoref{fig:normalisation-dist-other-lfw-vs-tinyface} provides an example for the ``Other'' data calibration variant, again for the QA algorithm CR-FIQA(L).
It shows that the QS distributions differ substantially for the two datasets, leading to a QS normalisation boundary calibration that doesn't fit the QSs on the other dataset well.
The mismatch would be even more severe if the ``Proportional'' calibration were used instead of the ``MinMax'' calibration,
since the densest QS concentrations differ (besides just the minima and maxima).

Actual EDC curve examples for the raw and the normalised QSs are depicted in \autoref{fig:normalisation-edc-lfw-same-vs-other}, for LFW with ``MinMax'' calibration on the ``Same'' and ``Other'' calibration data variants.
For the ``Same'' calibration data variant, the normalised QS curves exhibit a reasonably close fit to the raw QS curves.
However, normalised QS curves under the ``Other'' calibration data variant display substantial deficiencies towards higher discard fractions.
In this concrete example with LFW, this happens because the QSs on LFW tend to be higher than on TinyFace, as previously seen in \autoref{fig:normalisation-dist-other-lfw-vs-tinyface} for the QA algorithm CR-FIQA(L).

While the number of experiment configurations is too large to include all individual plots, more compact but complete results based on pAUC values for two discard ranges are listed in \autoref{tab:normalisation}.
From left to right, the columns show the pAUC discard range (0\% to either 20\% or 30\%), the calibration data variant (Same/Other/Combined), the calibration function (MinMax/Proportional), the dataset for the EDC evaluation (LFW/TinyFace),
the mean result of the QA algorithm result values (\ie{} across the row),
and then the individual results for the five QA algorithms.
Per configuration (table row), each QA algorithm (rightmost column set) has one raw/normalised QS EDC curve pair.
The corresponding result value is the area between these two curves divided by the pAUC value for the raw QS EDC curve, times 100.
This means that the result values are the percentages of the raw QS pAUC deviation caused by the normalisation, lower being better.

The results in \autoref{tab:normalisation} show that the configurations using the ``Other'' calibration data variant are substantially worse than the ones using the ``Same'' or ``Combined'' calibration data variants.
The differences between the ``Same'' and ``Combined'' variants are comparatively minor,
especially with the ``MinMax'' calibration function,
which demonstrates that a more general calibration (``Combined'') can be competitive with a best-case calibration (``Same'').
The ``Proportional'' calibration function does not appear to have a clear advantage over the simpler ``MinMax'' calibration function for ``Same''/``Combined''.
But it does have a clear disadvantage for ``Other'', since it is by design strongly affected by the distribution of the raw calibration QSs,
which becomes more apparent when higher pAUC discard limits are examined (here 30\% instead of 20\%).

In conclusion,
it may be preferable to evaluate different QA algorithms without normalisation first, before evaluating and selecting the best normalisation setup in a separate step,
since integer range normalisation is a strict reduction in precision that can substantially alter EDC curves depending on the approach.
A simple ``MinMax'' calibration can provide better results than a more QS-distribution-aware calibration such as ``Proportional''.
While a more expansive investigation of QS normalisation approaches is outside the scope of this EDC-centric paper, it could serve as a suitable topic for future work.

\section{Ranking stability with real data}
\label{sec:stability-real}

In this section we examine the ``stability'' of QA algorithm rankings resulting from the FNM-EDC pAUC values across different configurations, with real data as described in \autoref{sec:real-setup}.
Stability here refers to the similarity of the relative rankings described in \autoref{sec:pauc} for different FNM-EDC starting errors and pAUC ranges,
which can show us to which extent the rankings of QA algorithms can change with respect to these settings alone.

\subsection{EDC configuration ranges}
\label{sec:stability-edc-config}

The stability of the FNM-EDC-pAUC-based rankings is examined primarily using the real face image data across a grid of two parameter configurations:
\begin{itemize}
\item Starting error (\ie{} FNMR at 0 discard fraction): Range $[0.01,0.10]$ with a $0.01$ step (10 steps).
\item pAUC discard limit (\ie{} the maximum of the pAUC range): Range $[0.01,0.20]$ with a $0.01$ step (20 steps).
\\The pAUC range minimum is always 0.
\end{itemize}
This means that rankings are computed for $10\cdot 20 = 200$ EDC configurations.
Although greater ranges can be examined, \eg{} pAUC discard limit range $[0.01,0.99]$,
the ones used here were selected to approximate values one might find in the FIQA literature \cite{Schlett-FIQA-LiteratureSurvey-CSUR-2021},
and which are operationally relevant.
Note that if a biometric system in operation would discard 20\% of captured samples due to poor quality, the operation would not be considered successful.

Rankings are additionally examined for the fingerprint setup,
but as previously discussed in \autoref{sec:edc} this data happens to only allow for a more restricted starting error selection.
Here the six starting errors\footnote{Approximate values rounded to two digits.} 0.19, 0.31, 0.41, 0.50, 0.61, 0.70 are used.
The pAUC discard limit range is slightly extended to $[0.01,0.30]$, still with a $0.01$ step, yielding 30 steps and thus $6\cdot 30 = 180$ EDC configurations in total.
Although the larger values may no longer be considered operationally relevant, this part of the analysis can demonstrate whether some conclusions may even apply to more extreme configurations.

\subsection{Stability analysis}

\begin{figure}
\centering
\begin{tabular}{c}
LFW \\
\includegraphics[width=\linewidth]{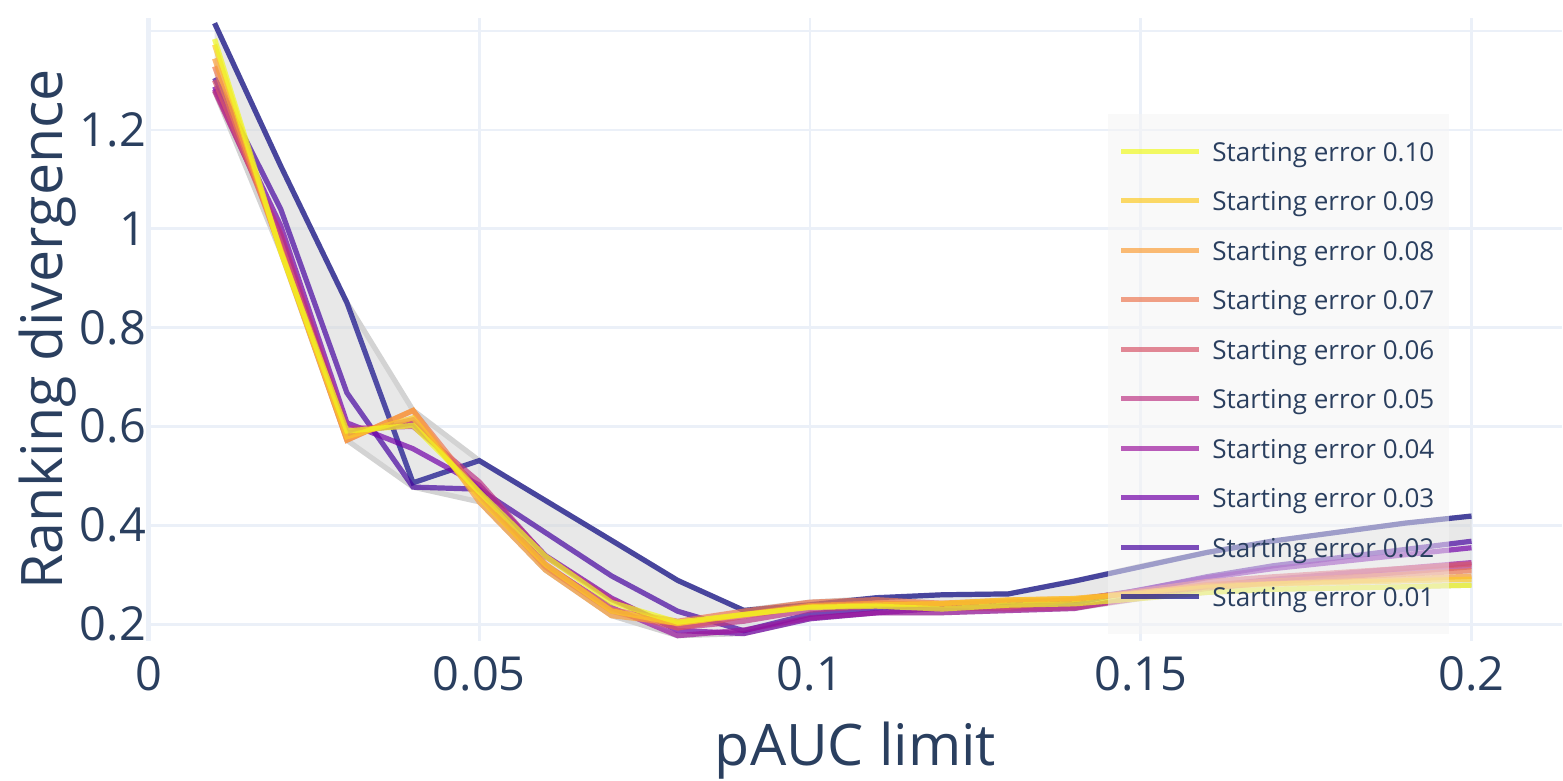} \\
TinyFace \\
\includegraphics[width=\linewidth]{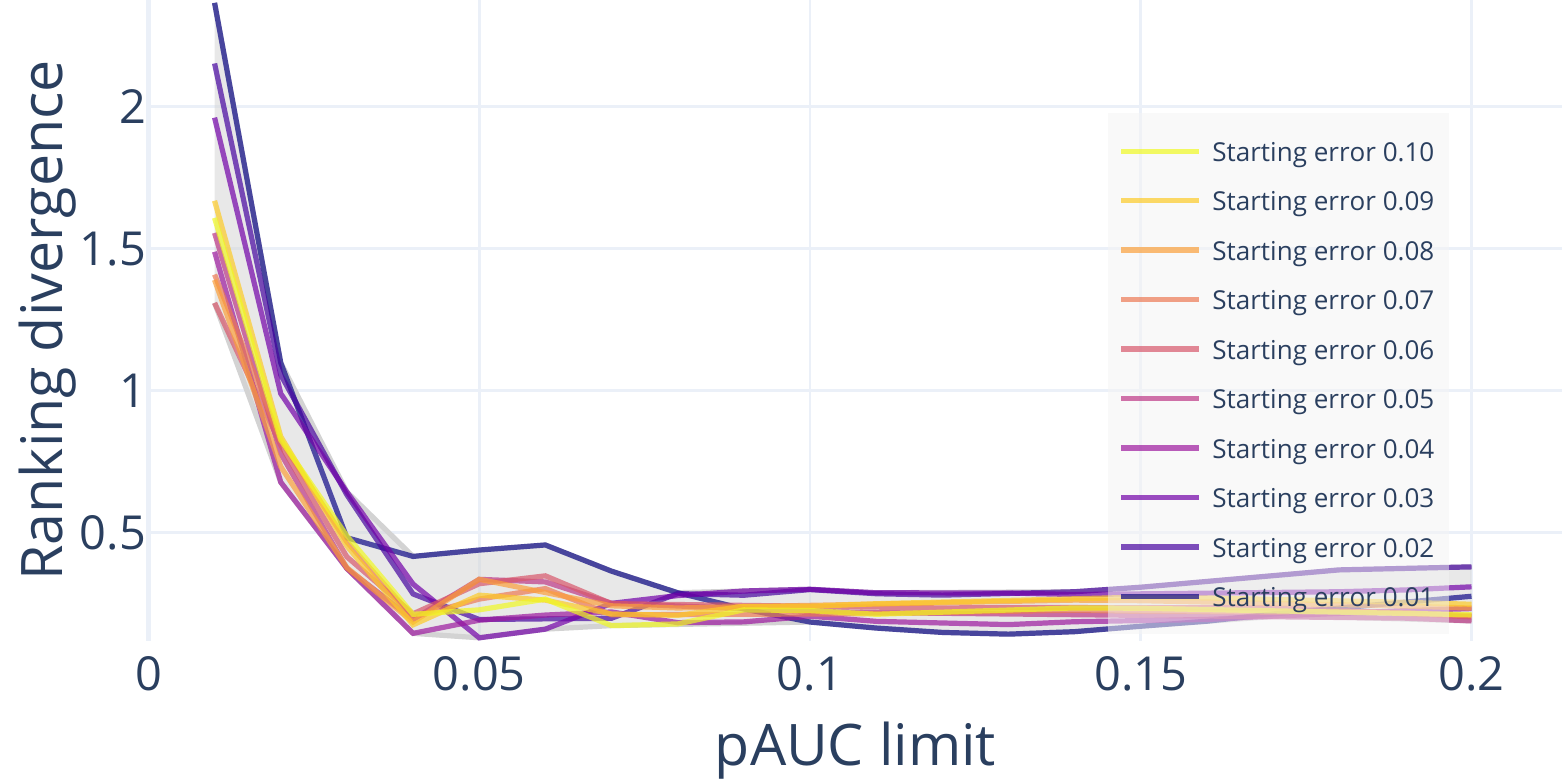}
\end{tabular}
\caption{\label{fig:stability-plots-real-data} EDC rankings \vs{} mean EDC ranking for real face image data. Curves correspond to the starting error 0.01 (dark purple) to 0.10 (bright yellow), with the pAUC limit varying across the X-axis. The Y-axis shows the difference of the ranking to the mean ranking (lower is better), times the QA algorithm count (five).}
\vspace{-1em}
\end{figure}

\begin{table}
\caption{\label{tab:stability-statistics-real-data}EDC ranking placement statistics for real face image data.}
\centering
\setlength{\tabcolsep}{3pt}
\textbf{LFW}\\
\begin{tabular}{r|rrr|rrr}
\theader{QA algorithm} & \theader{Span} & \theader{Best (Min)} & \theader{Worst (Max)} & \theader{Median} & \theader{Mean} & \theader{Std.dev.} \\
\hline
CR-FIQA(L) & 0.73 & 1.00 & 1.73 & 1.32 & 1.35 & 0.21 \\
CR-FIQA(S) & 2.38 & 2.06 & 4.44 & 3.86 & 3.82 & 0.59 \\
MagFace    & 0.01 & 1.00 & 1.01 & 1.00 & 1.00 & 0.00 \\
PCNet      & 2.19 & 2.81 & 5.00 & 3.27 & 3.65 & 1.00 \\
SER-FIQ    & 2.81 & 2.19 & 5.00 & 5.00 & 4.72 & 0.87 \\
\end{tabular}
\\\textbf{TinyFace}\\
\begin{tabular}{r|rrr|rrr}
\theader{QA algorithm} & \theader{Span} & \theader{Best (Min)} & \theader{Worst (Max)} & \theader{Median} & \theader{Mean} & \theader{Std.dev.} \\
\hline
CR-FIQA(L) & 1.40 & 1.00 & 2.40 & 1.10 & 1.23 & 0.33 \\
CR-FIQA(S) & 0.44 & 4.56 & 5.00 & 4.92 & 4.89 & 0.15 \\
MagFace    & 3.74 & 1.26 & 5.00 & 2.59 & 2.52 & 0.57 \\
PCNet      & 4.00 & 1.00 & 5.00 & 5.00 & 4.56 & 1.28 \\
SER-FIQ    & 2.25 & 1.00 & 3.25 & 1.00 & 1.08 & 0.42 \\
\end{tabular}
\vspace{-2em}
\end{table}

\begin{figure}
\centering
\begin{tabular}{c}
FVC2006-DB2\_A \\
\includegraphics[width=\linewidth]{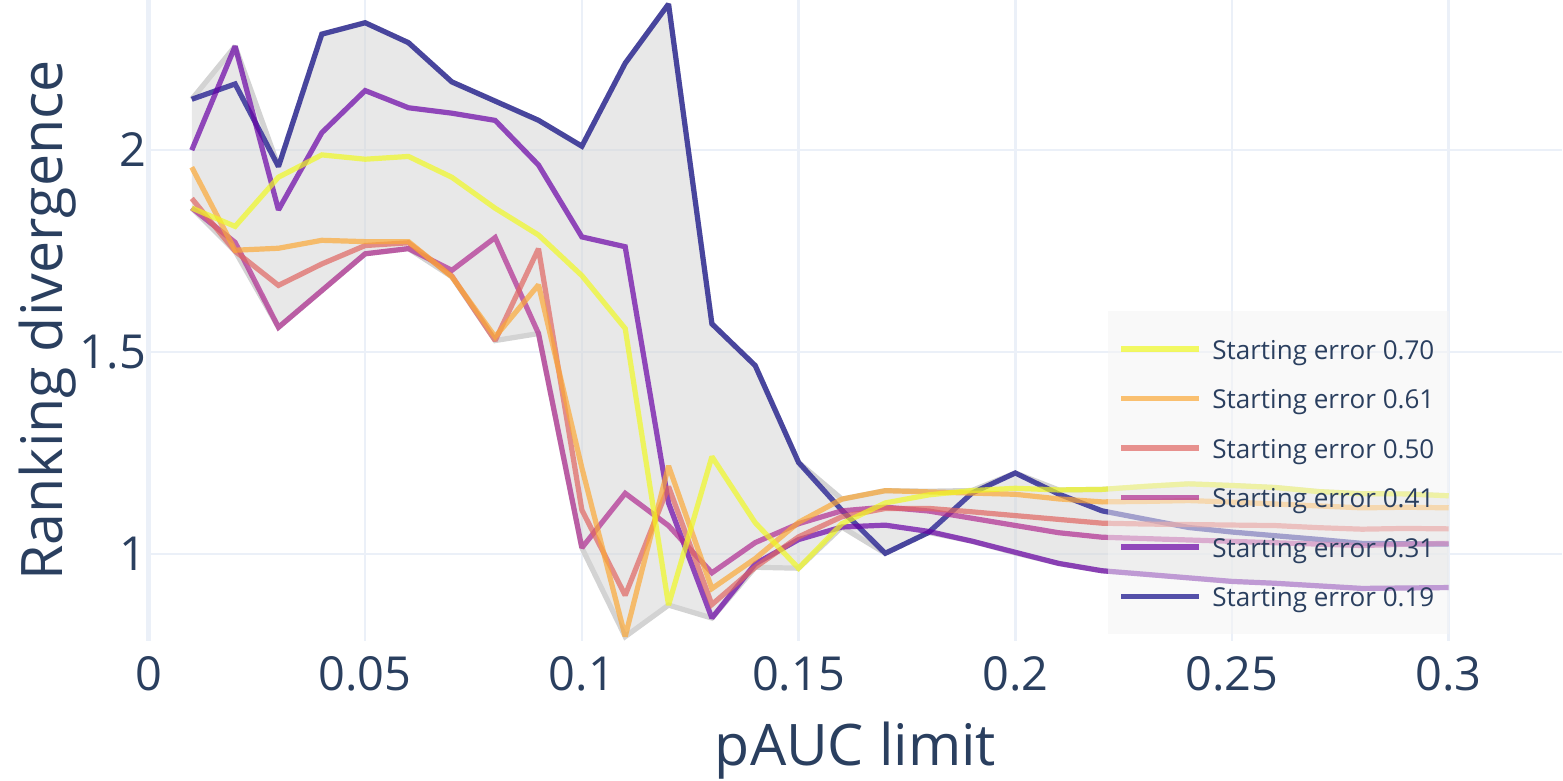}
\end{tabular}
\caption{\label{fig:stability-plots-real-data-fingerprint} EDC rankings \vs{} mean EDC ranking for real fingerprint data, analogous to \autoref{fig:stability-plots-real-data}.}
\vspace{-1em}
\end{figure}

\autoref{fig:stability-plots-real-data} depicts the ``stability'' of the FNM-EDC-pAUC-based face image QA rankings for the 200 FNM-EDC-pAUC configurations.
The X-axis shows the pAUC limit configurations (20 steps), while the different curves correspond to the 10 starting error configurations.
The Y-axis shows the instability in terms of the rankings' divergence from the mean ranking, times the number of QA algorithms (here five), lower being better (more stable).
These instability values (one data point) are computed via \autoref{eq:stability-real},

\begin{equation}\label{eq:stability-real}
RankingDivergence = \Sigma^n_i |p_i-\bar{p_i}|
\end{equation}

$n$ being the QA algorithm count (five), $p_i$ being one QA algorithm's placement in the ranking, and $\bar{p_i}$ being the mean placement of the QA algorithm among all 200 rankings.
The rankings are the ``relative rankings'' previously described at the end of \autoref{sec:pauc},
meaning that a QA algorithm placement value in a ranking is the min-max normalised pAUC value for the QA algorithm's FNM-EDC curve.

Be aware that this comparison against the mean ranking is biased insofar that rankings with configurations closer to the mean configuration\footnote{Here pAUC discard limit 0.10 and starting error 0.05.} will inherently be more similar to this mean ranking, thus yielding lower instability values.
It is however possible to examine whether certain configurations exhibit instability values that cannot be explained by this bias alone.
The results illustrated in \autoref{fig:stability-plots-real-data} thus primarily indicate that the ranking divergence can increase substantially at lower pAUC limits,
but not at higher pAUC limits, irrespective of the starting error choice (\ie{} higher pAUC limits diminish the effect of the starting error choice on the ranking divergence).
Based on these results it appears that higher pAUC limits should be preferred.
Please note that we refrain from recommending more specific value bounds for these parameters,
since the stability is ultimately also dependent on the dataset and the used recognition/QA algorithms, so that concrete stable configuration values seen here may not translate precisely to other evaluation setups.

\autoref{tab:stability-statistics-real-data} lists ranking statistics per QA algorithm across the 200 rankings.
The ``Median''/``Mean'' columns show the median/mean ranking placements in the range $[1,5]$, lower being better and five being the QA algorithm count,
with ``Std.dev.'' showing the placement standard deviation in the range $[0,5/2]$\footnote{As described previously, the originally computed rankings are in the range $[0,1]$. The values in the table are adjusted by the QA algorithm count, five, to improve the interpretability.}.
FIQA algorithm evaluation is not the focus of this paper, but we can observe that the mean and median results here show CR-FIQA(L) and MagFace as more effective for both datasets. Both use a similar network architecture (ResNet100) and, perhaps more importantly, training data (MS1MV2 \cite{Deng-ArcFace-IEEE-CVPR-2019}), but their training procedures differ.
SER-FIQ was the best algorithm for TinyFace but the worst for LFW.
MagFace also appears to be especially effective for LFW relative to the other algorithms.

The ``Best''/``Worst'' placement columns in \autoref{tab:stability-statistics-real-data} with range $[1,5]$ provide additional insight regarding ranking stability,
with ``Span'' being the worst placement minus the best placement value (range $[0,5-1]$):
PCNet on TinyFace is an extreme example with the maximum span value four,
meaning that there is one EDC configuration for which PCNet appears as the best algorithm in the ranking,
and one EDC configuration where it appears as the worst algorithm.
More generally, the placement span for most of the QA algorithms in both datasets is above one.
This means that it is possible to optimise the placement for one QA algorithm by selecting a specific EDC configuration for one dataset.
Restricting the EDC parameters to more stable configurations means that such placement optimisation becomes more restricted,
even if only one dataset is considered in isolation.

As \autoref{fig:stability-plots-real-data-fingerprint} shows, the analogous analysis using the fingerprint setup supports the face image results, using more extreme configuration values and a different modality.

\section{Ranking stability with synthetic data}
\label{sec:stability-synthetic}

A noteworthy limitation of the stability analysis in the previous section is that there was no expected (\ie{} already known) ranking of the real QA algorithms,
which is why we compared rankings against the mean ranking.
This limitation can be circumvented through fully synthetic EDC input data with a predefined ranking.

\subsection{Synthetic data setup}

\begin{figure}
\centering
\begin{tabular}{c}
Variant 1 \\
\includegraphics[width=0.99\linewidth]{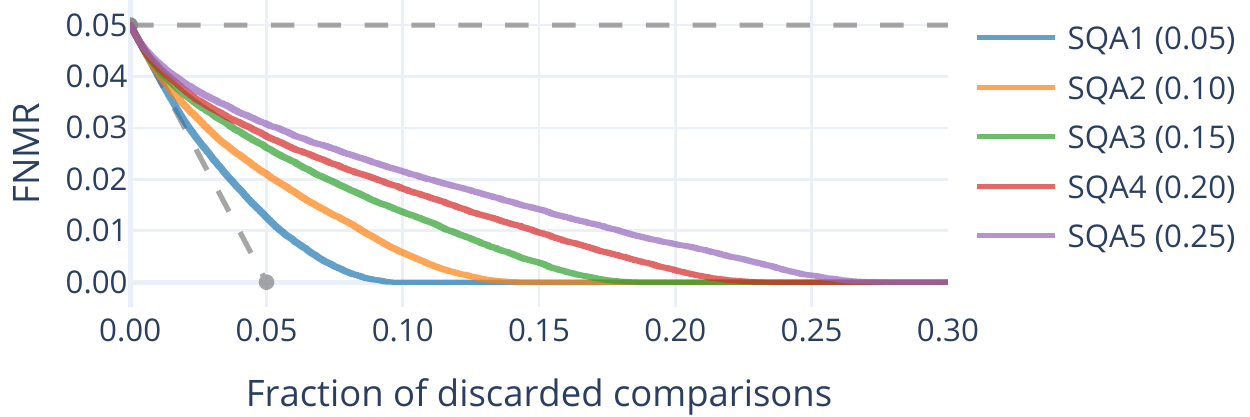} \\
Variant 2 \\
\includegraphics[width=0.99\linewidth]{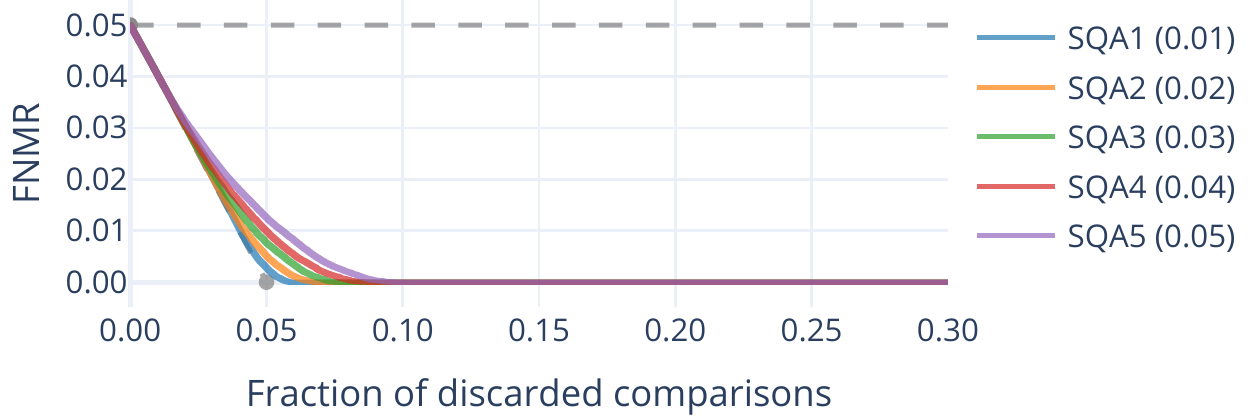}
\end{tabular}
\caption{\label{fig:stability-analysis-dataset-edc-examples-synthetic-data} EDC plot examples for the synthetic data.}
\end{figure}

Here ``fully synthetic'' means that no biometric samples are involved, only synthetic scores.
First, synthetic sample ``utility scores'' are randomly generated\footnote{$[-1,+1]$ range, but this range is not functionally important.},
each utility score representing one imaginary sample.
The set of utility scores is hidden to the QA algorithm ranking computation, insofar that it is only used to derive synthetic floating point QSs and mated CSs,
but it is not directly used as an input for the FNM-EDC evaluation.
For these experiments $50,000$ synthetic subjects with five samples each are considered, which means there are $50,000\cdot 5=250,000$ generated sample utility scores and $50,000\cdot {{5}\choose{2}}=500,000$ mated pairs.

Synthetic mated CSs are the pairwise minima of the utility scores, effectively assuming that the CS for a mated pair exclusively depends on the sample with lower utility.
Synthetic sample QSs are the utility scores plus/minus some random offset.
Different synthetic QA algorithms are defined simply by scaling this random offset for sample QS generation.
The random distributions are uniform for clarity.
As described in \autoref{sec:edc}, a pairwise QS is the minimum of the pair's sample QSs,
so without the random offset the pairwise QSs would be identical to the CSs in this synthetic setup.

\begin{figure}
	\centering
	\begin{tabular}{c}
		Variant 1 \\
		\includegraphics[width=\linewidth]{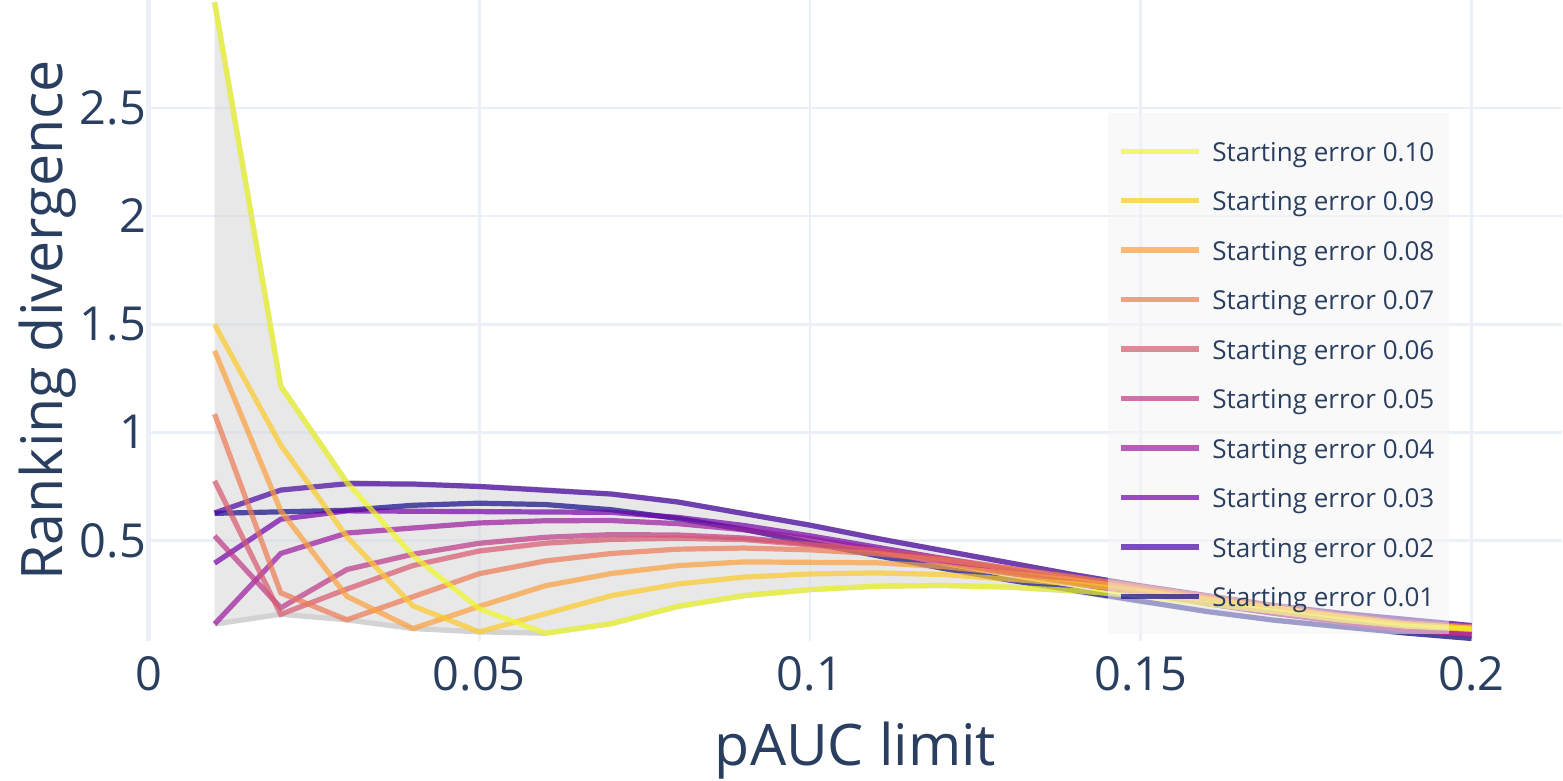} \\
		Variant 2 \\
		\includegraphics[width=\linewidth]{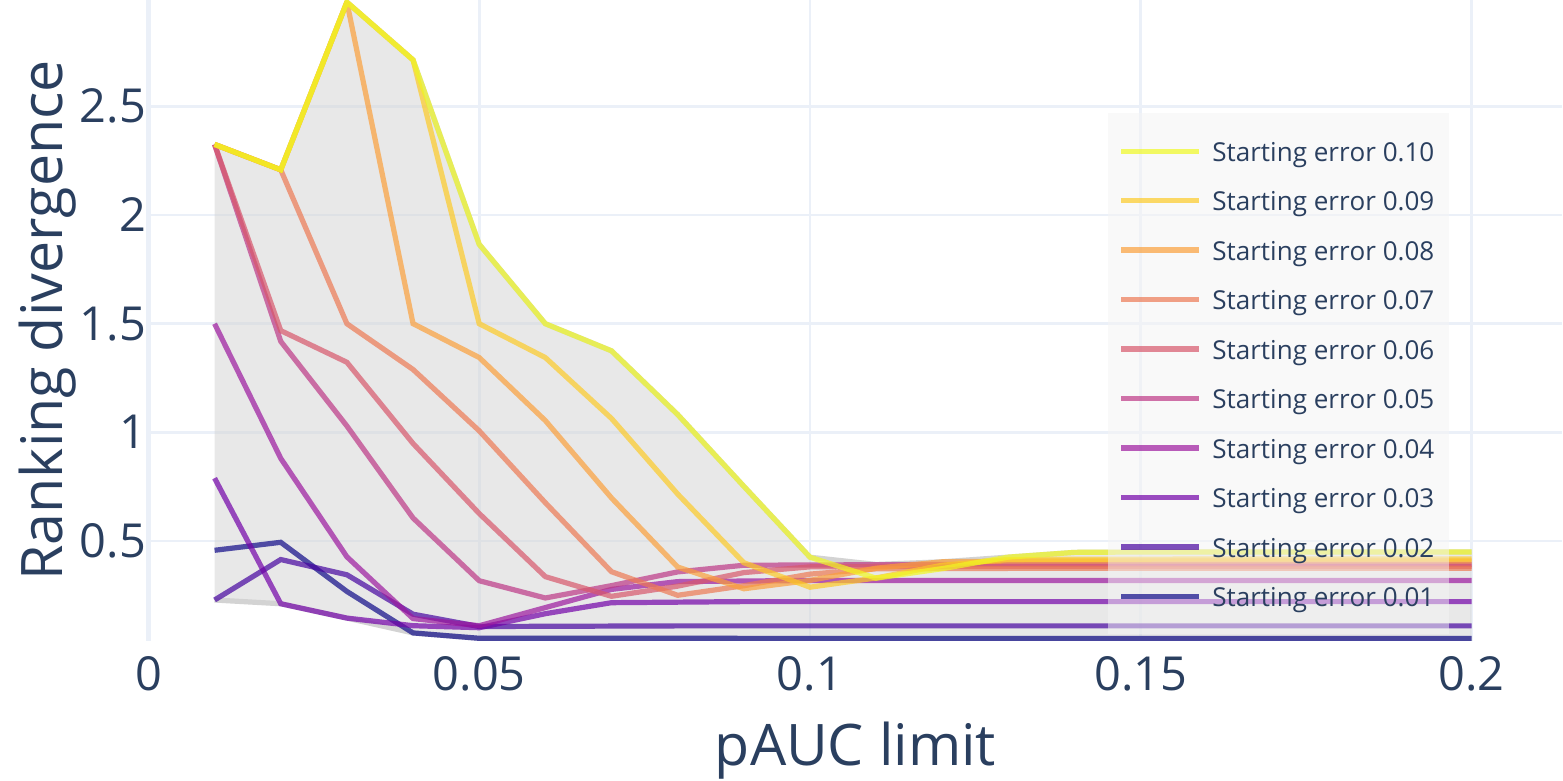}
	\end{tabular}
	\caption{\label{fig:stability-plots-synthetic-data} EDC rankings \vs{} expected ranking for synthetic data. Curves correspond to the starting error 0.01 (dark purple) to 0.10 (bright yellow), with the pAUC limit varying across the X-axis. The Y-axis shows the difference of the actual ranking to expected ranking (lower is better), times the QA algorithm count (five).}
\end{figure}

\begin{table}
	\caption{\label{tab:stability-statistics-adjusted-synthetic-data}Ranking stability statistics for synthetic data.}
	\centering
	\setlength{\tabcolsep}{1.8pt}
	\textbf{Variant 1}\\
	\begin{tabular}{r|rrr|rrr}
		\theader{QA algorithm} & \theader{Span} & \theader{Best (Min)} & \theader{Worst (Max)} & \theader{Median} & \theader{Mean} & \theader{Std.dev.} \\
		\hline
		SQA1 (0.05) & 1.57 & 1.00 & 2.57 & 1.00 & 1.01 & 0.14 \\
		SQA2 (0.10) & 4.00 & 1.00 & 5.00 & 2.30 & 2.31 & 0.61 \\
		SQA3 (0.15) & 3.99 & 1.00 & 4.99 & 3.55 & 3.47 & 0.71 \\
		SQA4 (0.20) & 3.84 & 1.00 & 4.84 & 4.38 & 4.29 & 0.64 \\
		SQA5 (0.25) & 2.40 & 2.60 & 5.00 & 5.00 & 4.99 & 0.21 \\
	\end{tabular}
	\\\textbf{Variant 2}\\
	\begin{tabular}{r|rrr|rrr}
		\theader{QA algorithm} & \theader{Span} & \theader{Best (Min)} & \theader{Worst (Max)} & \theader{Median} & \theader{Mean} & \theader{Std.dev.} \\
		\hline
		SQA1 (0.01) & 4.00 & 1.00 & 5.00 & 1.00 & 1.24 & 1.11 \\
		SQA2 (0.02) & 4.00 & 1.00 & 5.00 & 1.54 & 1.63 & 0.67 \\
		SQA3 (0.03) & 4.00 & 1.00 & 5.00 & 2.49 & 2.57 & 1.01 \\
		SQA4 (0.04) & 3.69 & 1.00 & 4.69 & 3.58 & 3.41 & 1.02 \\
		SQA5 (0.05) & 3.59 & 1.41 & 5.00 & 5.00 & 4.85 & 0.71 \\
	\end{tabular}
\vspace{-1em}
\end{table}

Two variants with different QS generation offset scaling values for the five synthetic QA algorithms are used in the following:
\begin{itemize}
\item Variant 1: $0.05,0.10,0.15,0.20,0.25$
\item Variant 2: $0.01,0.02,0.03,0.04,0.05$
\end{itemize}

The expected ranking for these synthetic QA algorithms corresponds to the order of the scaling values.
\Eg{} the lowest scaling value results in sample QSs closest to the utility scores, which in turn should result in the best placement within the computed ranking if the evaluation works as intended.

\autoref{fig:stability-analysis-dataset-edc-examples-synthetic-data} shows EDC plot examples for both synthetic variants,
whereby the labels of the synthetic QA (``SQA'') algorithms denote the expected ranking and the random offset scaling.
For example, ``SQA3 (0.15)'' has expected ranking 3 with random offset scaling 0.15.
The EDC curves differ substantially from real data curves (\eg{} \autoref{fig:edc-plot-example}),
the most relevant difference being that the expected ranking of the five synthetic QA algorithms is readily apparent by looking at the curves.
This indicates that the quantified pAUC rankings should also be able to reflect the expected ranking.

\subsection{Stability analysis}

The same 200 EDC pAUC discard limit and starting error configurations
used for the stability analysis on real face image data, defined in \autoref{sec:stability-edc-config},
are also used here.
In contrast to the real data analysis,
this analysis with synthetic data compares the 200 rankings against the expected ranking implied by the order of the synthetic QS generation offset scaling values,
instead of the mean ranking.
The instability is plotted in \autoref{fig:stability-plots-synthetic-data}, similar to the plot for real data in \autoref{fig:stability-plots-real-data}, but comparing against expected rankings in \autoref{eq:stability-synth},

\begin{equation}\label{eq:stability-synth}
RankingDivergence = \Sigma^n_i |p_i-e_i|
\end{equation}

$n$ being the QA algorithm count (here five), $p_i$ being one QA algorithm's placement in the ranking, and $e_i$ being the expected placement of the QA algorithm.
The expected placement value is the min-max normalised QS generation offset scaling value, so the values are $0.00, 0.25, 0.50, 0.75, 1.00$ for both variant 1 and 2.

Analogous to the observations in the real data analysis,
low pAUC limits can lead to rankings that deviate strongly from the expected rankings,
in contrast to higher pAUC limits for which the ranking divergence is comparatively small irrespective of the starting error choice.
At low pAUC limits, note that there is no specific starting error that leads to the lowest ranking divergence across all cases (\ie{} dataset and QA algorithm setups), which can be observed both for the synthetic data (\autoref{fig:stability-plots-synthetic-data}) and the real data (\autoref{fig:stability-plots-real-data}, \autoref{fig:stability-plots-real-data-fingerprint}).

\autoref{tab:stability-statistics-adjusted-synthetic-data} lists statistics
similar to \autoref{tab:stability-statistics-real-data} from the real data analysis.
The ``Median'' and ``Mean'' rankings approximate the expected rankings 1 to 5 from top to bottom,
and the ``Span'' column values highlight that all of the synthetic FIQA algorithms' individual rankings could be optimised by selecting specific pAUC limit/starting error configurations,
as long as those configurations are not constrained to higher pAUC limits.

In summary,
the main advantage of this synthetic analysis over the real data analysis is that the stability is tested relative to expected rankings, which are unknown for real algorithms.
The synthetic analysis supports the conclusion that higher (but still operationally relevant) pAUC limits should be preferred,
while the choice of the starting error becomes less important with higher pAUC limits.

\section{Other evaluation approaches}
\label{sec:other-approaches}

Experiments in this paper focus on the (FNM-)EDC for multiple reasons:
\begin{enumerate}
\item As noted in the introduction, the EDC is being standardised in the next edition of ISO/IEC 29794-1.
\item Evaluating the QA algorithm performance via FNM-EDC plots with one recognition system and the minimum as the pairwise QS function reflects the common approach employed in the contemporary literature.
\item EDC plots in general show directly how a biometric recognition error value changes as samples/comparisons are discarded based on sample QSs.
In contrast, other non-EDC approaches presented in the remainder of this section consider only parts of these operationally relevant biometric concepts.
\end{enumerate}

\begin{figure}
\centering
\begin{tabular}{c}
TinyFace \\
\includegraphics[width=0.97\linewidth]{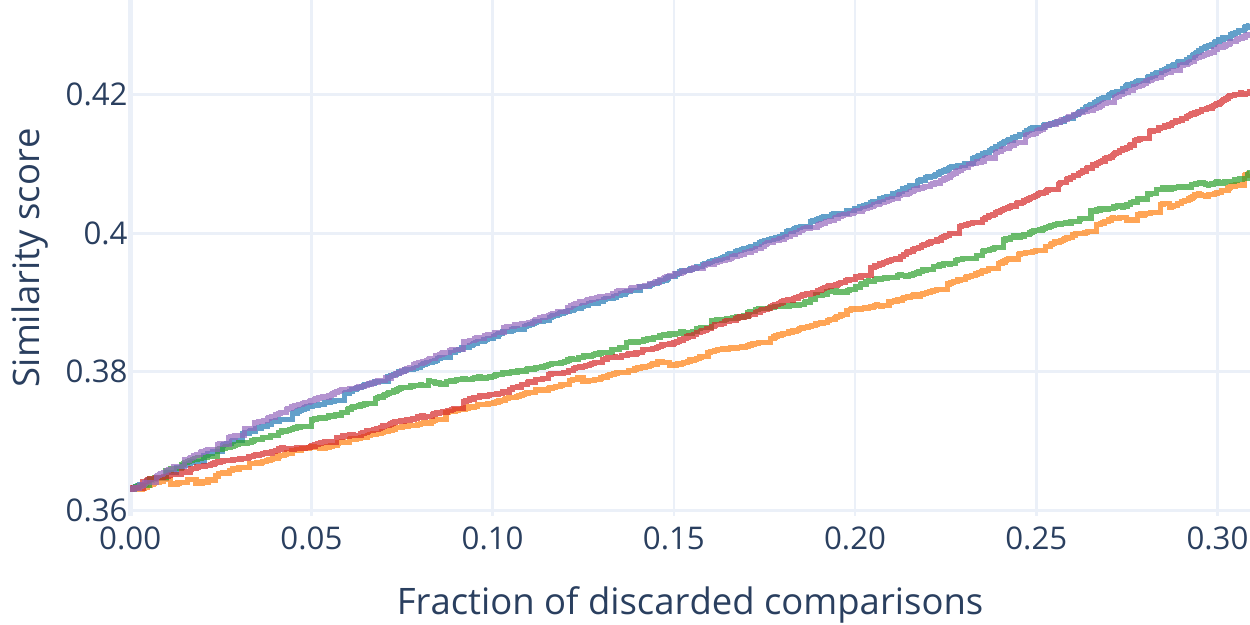} \\
\includegraphics[width=0.97\linewidth]{img/real-data/Real-FIQA-legend} \\
\end{tabular}
\caption{\label{fig:cs-dc-example} CS-DC example plot showing mean mated similarity scores.}
\vspace{-1em}
\end{figure}

One non-EDC approach is to plot the mean comparison scores instead of error values.
In this section this approach is called ``CS-DC'' for ``Comparison Score versus Discard Characteristic''.
For the sake of simplicity all CSs are assumed to be similarity scores in this section.
\autoref{fig:cs-dc-example} shows an example.
For mated comparisons higher/increasing similarity scores are better,
and for non-mated comparisons lower/decreasing similarity scores are better.
Although it is technically possible to use both mated and non-mated comparisons within the same curve by using similarity scores for one and dissimilarity scores for the other (e.g. by negating the scores for one type),
this can distort the interpretability of the results,
since the mated and non-mated comparisons should inherently be represented by different CS distributions.
One CS-DC curve should therefore use either only mated or non-mated comparisons.
This is analogous to the difference between FNM-EDC curves (False Non-Match error using mated comparisons) and FM-EDC curves (False Match error using non-mated comparisons).
The difference to FNM-EDC and FM-EDC curves is that the CS-DC curves do not consider a CS threshold that divides the comparisons into error and non-error cases.

\begin{figure}
\centering
\begin{tabular}{c}
TinyFace \\
\includegraphics[width=0.97\linewidth]{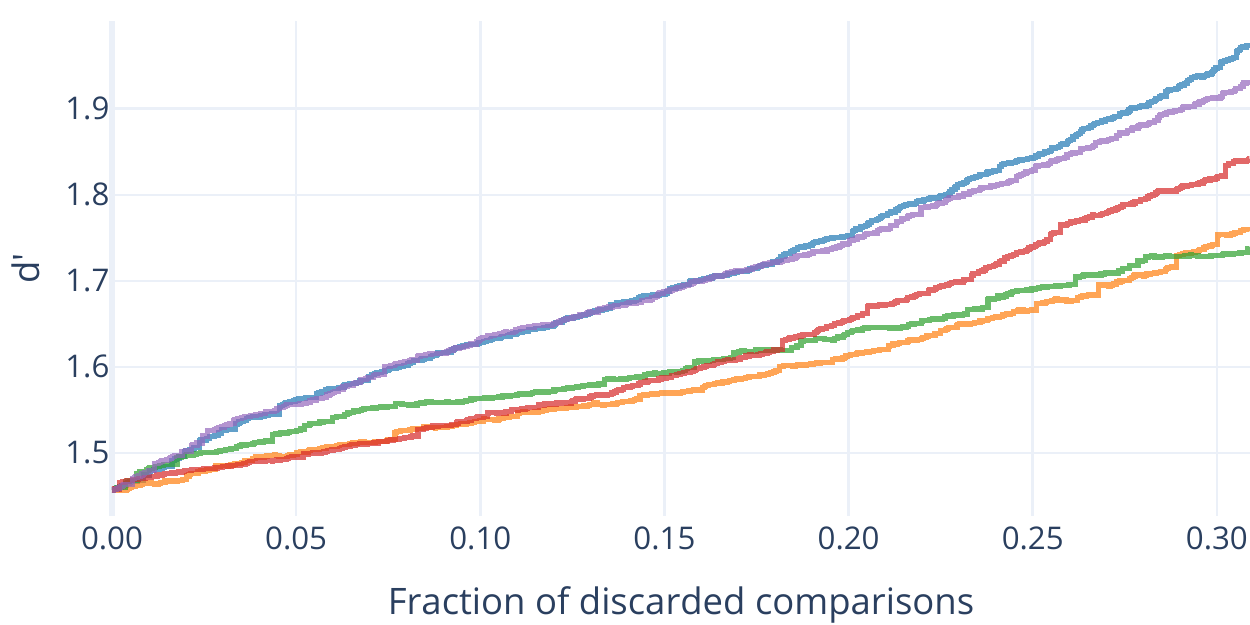} \\
\includegraphics[width=0.97\linewidth]{img/real-data/Real-FIQA-legend} \\
\end{tabular}
\caption{\label{fig:dprime-dc-example} $d'$-DC example plot.}
\vspace{-1em}
\end{figure}

Another non-EDC approach is the ``$d'$ versus discard characteristic'' proposed by \markAuthor{Henniger \etal{}} \cite{Henniger-QA-UtilityBasedEvaluation-BIOSIG-2022},
abbreviated as ``$d'$-DC'' in this section.
Similar to the EDC and the CS-DC, samples/comparisons are discarded based on QSs, which is plotted from left to right on the X-axis.
But as per \autoref{eq:dprime},
instead of an error value this approach plots $d'$,
which is the difference between the mean of the remaining non-mated dissimilarity scores ($\mu_n$) and the mean of the remaining mated dissimilarity scores ($\mu_m$),
normalised by the corresponding standard deviation values ($\sigma$).
\begin{equation}\label{eq:dprime}
d' = \frac{\mu_n - \mu_m}{\sqrt{\sigma^2_n+\sigma^2_m}}
\end{equation}
\Ie{} for $d'$ higher/increasing values are better.
See \autoref{fig:dprime-dc-example} for an example.
The advantage of the $d'$-DC is that both mated and non-mated CSs are considered in the same curve,
but similar to the CS-DC this does not consider any CS threshold to divide comparisons into error and non-error cases.

\begin{figure}
\centering
\begin{tabular}{c}
TinyFace \\
\includegraphics[width=0.97\linewidth]{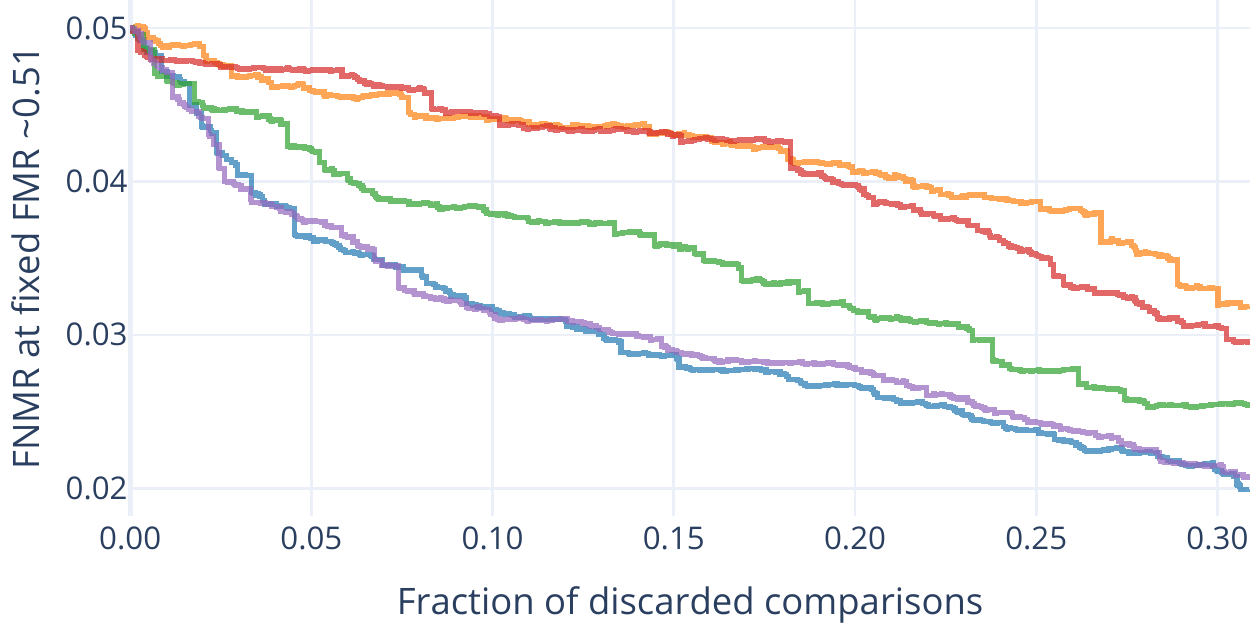} \\
\includegraphics[width=0.97\linewidth]{img/real-data/Real-FIQA-legend} \\
\end{tabular}
\caption{\label{fig:fc-edc-example} FC-EDC example plot.}
\vspace{-1em}
\end{figure}

Note that it is however also possible to consider both mated and non-mated CSs in a single EDC curve.
This can be done for instance by plotting the changing FNMR against an approximately fixed FMR (or vice versa) for a given set of mated and non-mated comparisons,
which is called ``FC-EDC'' for ``Fixed Counterpart EDC'' in this section.
\autoref{fig:fc-edc-example} shows an example.
In this paper the FNM-EDC was prioritized for experiments due to the aforementioned prevalence in the literature and due to current standardisation efforts,
but it can be argued that an FC-EDC is a more straightforward EDC type for operationally relevant evaluations,
since it considers both the FNMR and the FMR in a single plot.

Many of the general conclusions in this paper also apply to the just described CS-DC, $d'$-DC, and FC-EDC:
\begin{itemize}
\item pAUC (\autoref{sec:pauc}): pAUC values can be computed to consider a range of discard fractions.
\item Curve interpolation (\autoref{sec:interpolation}): Since only discrete numbers of samples/comparisons are discarded along the X-axis, it is advisable to use stepwise interpolation for all of these evaluation plot types.
\item Quality score normalisation (\autoref{sec:normalisation}): Deviations introduced by normalisation can naturally affect any evaluation. For plots this can have similar effects as in the shown FNM-EDC examples.
\item Ranking stability (\autoref{sec:stability-real} and \autoref{sec:stability-synthetic}): As especially the experiments with fully synthetic scores in \autoref{sec:stability-synthetic} should demonstrate, QA algorithm rankings based on pAUC values can in general differ substantially for different pAUC discard limits, and will likely stabilize relative to each other as higher pAUC discard limits are used.
This consideration may however be obviated if operational requirements clearly specify the evaluation parameters,
and the FC-EDC in particular allows for the concrete specification of a fixed FMR or FNMR.
\end{itemize}

Besides the various described curve-based evaluations,
it is possible to directly compute scalar values to rank QA algorithms by assessing the correlation between QSs and CSs in some way.
One way to do this is by computing the Pearson correlation coefficient between the CSs and the corresponding pairwise QSs (those \eg{} being the sample QS minimum, as previously described for typical EDC curves).
Another way proposed by \markAuthor{Henniger \etal{}} \cite{Henniger-QA-UtilityBasedEvaluation-BIOSIG-2022}
is to first compute sample ``utility'' scores similar to $d'$ (see the paper for details),
and then compute Root Mean Square Error values between these scores and the QSs.
This may not work well if the scores are not using an identical or at least comparable value range,
so we propose to instead compute the Pearson correlation coefficient between the sample utility scores and the QSs
(cf.\@ \cite{Olsen-FingerImageQuality-IETBiometrics-2016} with Spearman correlation).
All three of these correlation-based QA algorithm rankings can consider both mated and non-mated comparisons at once,
but they have the disadvantage that they do not consider the concept of discarding samples based on the QSs (although there may be use cases for which this is not relevant).
They also do not divide comparisons into error and non-error cases, similar to the CS-DC and the $d'$-DC,
which could however technically be done by \eg{} assessing the correlation between the pairwise QSs and error/non-error proxy values such as 0/1.

Yet another evaluation approach is the inspection of Detection Error Trade-off (DET) curves across different QS thresholds, which shows the FNMR and the corresponding FMR on the two axes.
This can either be done similar to EDC plots by discarding samples/comparisons for each QS threshold,
or the QS thresholds can be used to separate the samples/comparisons into multiple disjunct bins from which the curves are computed.
The former approach is also being standardised in the next edition of ISO/IEC 29794-1 as the ``DET versus discard method''.
While the EDC and the other previously described curve-based evaluation approaches can inherently consider all possible QS thresholds in one curve,
a disadvantage of this DET-based approach is that only a smaller selection of QS thresholds can be sensibly examined/visualized,
since each added QS threshold for each QA algorithm results in a new DET curve.

As shown in this section there are various possible QA algorithm evaluation approaches.
Naturally not every feasible variation is included here,
but, as discussed, the EDC variations already cover common operationally relevant considerations.

\section{Conclusions}
\label{sec:conclusions}

We discussed various general aspects of EDC QA algorithm evaluation,
as well as aspects of quantitative rankings based on pAUC values in particular,
with examples focusing on FNM-EDC configurations.
The primary findings can be summarised as follows:

\begin{enumerate}
\item Stepwise curve interpolation should be preferred for both for graphical plots and for pAUC value computation, to reflect the actual behaviour of the error with respect to the discard steps.
\item Relative rankings based on pAUC values can be used to show how close each QA algorithm is to the best/worst performing one.
\item If pAUC values are examined directly (instead of \eg{} the relative rankings), the interpretability can be improved by normalising them relative to the hard lower error limit defined by the ``area under theoretical best'', $max(0, Error-DiscardFraction)$, and the soft upper error limit defined by the starting error constant.
\item Normalising quality scores to an integer range such as $[0, 100]$ does naturally affect the EDC curves, and there are different normalisation approaches that also depend on the selected calibration data. It may therefore be preferable to evaluate different QA algorithms without normalisation, before evaluating and selecting the best normalisation setup in a separate step.
\item Despite the simplicity, simple min-max integer range normalisation of quality scores can be effective, while a normalisation proportional to calibration quality scores can allocate too much detail to higher quality scores, which is counterproductive since the differentiation between lower quality levels is more important.
\item Quantitative rankings based on pAUC values were shown to be less reliable for low pAUC discard limits, so higher operationally relevant discard limits should be preferred.
The starting error choice appeared to be less important at these higher discard limits.
\end{enumerate}

We are recommending that researchers and people involved in standardisation take the considerations of this work into account.

\ifCLASSOPTIONcompsoc
\section*{Acknowledgments}
\else
\section*{Acknowledgment}
\fi

This research work has been funded by the German Federal Ministry of Education and Research and the Hessian Ministry of Higher Education, Research, Science and the Arts within their joint support of the National Research Center for Applied Cybersecurity ATHENE.
This project has received funding from the European Union’s Horizon 2020 research and innovation programme under grant agreement No 883356.
This text reflects only the author’s views and the Commission is not liable for any use that may be made of the information contained therein.

We thank Sven Utcke for bringing the idea behind what this paper calls the ``FC-EDC'' to our attention.

\section*{References}
\AtNextBibliography{\small}
\printbibliography[heading=none]

\end{document}